\documentclass[default,iicol,sn-mathphys,pdflatex]{sn-jnl}% Default with double column layout

\jyear{2021}%

\theoremstyle{thmstyleone}%
\theoremstyle{thmstyletwo}%

\theoremstyle{thmstylethree}%

\definecolor{revise_text}{rgb}{0,0,0}

\begin{document}

\title[Article Title]{Instance-dependent Label Distribution Estimation for Learning with Label Noise}


\author[1]{\fnm{Zehui} \sur{Liao}}\email{merrical@mail.nwpu.edu.cn}
% ORCID: 0000-0002-8475-5819
\author[1]{\fnm{Shishuai} \sur{Hu}}\email{sshu@mail.nwpu.edu.cn}
% ORCID: 0000-0002-7314-6647
\author[2]{\fnm{Yutong} \sur{Xie}}\email{yutong.xie678@gmail.com}
% ORCID: 0000-0002-6644-1250
\author*[1]{\fnm{Yong} \sur{Xia}}\email{yxia@nwpu.edu.cn}
% ORCID: 0000-0001-9273-2847
\affil[1]{School of Computer Science, Northwestern Polytechnical University, Xi’an 710072, China}
\affil[2]{Australian Insititute for Machine Learning, The University of Adelaide, Adelaide, SA 5000, Australia}

\abstract{
Noise transition matrix estimation is a promising approach for learning with label noise. It can infer clean posterior probabilities, known as Label Distribution (LD), based on noisy ones and reduce the impact of noisy labels. However, this estimation is challenging, since the ground truth labels are not always available.
Most existing methods estimate a global noise transition matrix using either correctly labeled samples (anchor points) or detected reliable samples (pseudo anchor points). These methods heavily rely on the existence of anchor points or the quality of pseudo ones, and the global noise transition matrix can hardly provide accurate label transition information for each sample, since the label noise in real applications is mostly instance-dependent.
To address these challenges, we propose an Instance-dependent Label Distribution Estimation (ILDE) method to learn from noisy labels for image classification. 
\textcolor{revise_text}{The method's workflow has three major steps.
First, we estimate each sample's noisy posterior probability, supervised by noisy labels.
Second, since mislabeling probability closely correlates with inter-class correlation, we compute the inter-class correlation matrix to estimate the noise transition matrix, bypassing the need for (pseudo) anchor points. 
Moreover, for a precise approximation of the instance-dependent noise transition matrix, we calculate the inter-class correlation matrix using only mini-batch samples rather than the entire training dataset. 
Third, we transform the noisy posterior probability into instance-dependent LD by multiplying it with the estimated noise transition matrix, using the resulting LD for enhanced supervision to prevent DCNNs from memorizing noisy labels.}
The proposed ILDE method has been evaluated against several state-of-the-art methods on two synthetic and \textcolor{revise_text}{three} real-world noisy datasets. 
Our results indicate that the proposed ILDE method outperforms all competing methods, no matter whether the noise is synthetic or real noise.
}

\keywords{Label Noise, Label Distribution, Noise Transition Matrix, Inter-class Correlation}

\maketitle

\begin{figure*}[h]
\centering
\includegraphics[width=2.0\columnwidth]{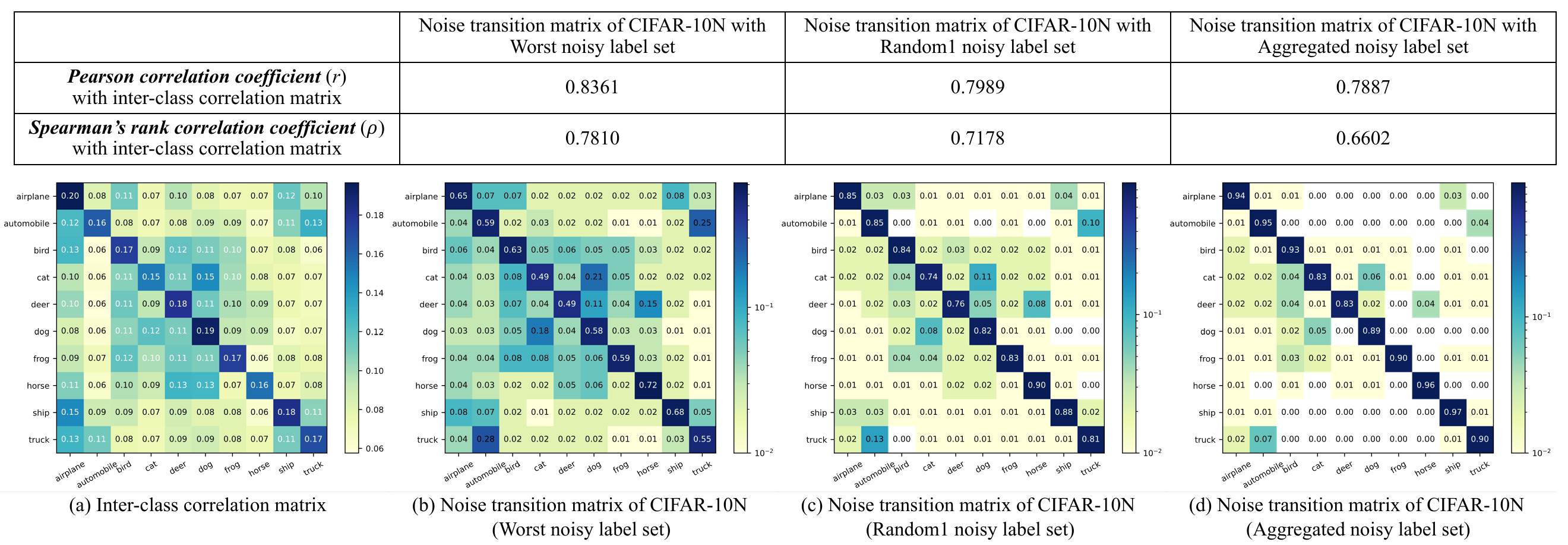}
\caption{
\textcolor{revise_text}{Inter-class correlation matrix and three noise transition matrices derived from the `Worst', `Random1', `Aggregated' noisy label sets of CIFAR-10N, with the three color bars on the right-side log-norm transformed for clarity. The noise transition matrices are correlated with the class correlation matrix, with the Pearson and Spearman’s rank correlation coefficients calculated and displayed, highlighting the relationship between class similarity and mislabeling tendencies.}
}
\label{fig: intro}
\end{figure*}

\section{Introduction}
\label{sec:intro}
Deep convolutional neural networks (DCNNs) rely heavily on accurately labeled training data for their success. However, the collected datasets often contain label noise due to the complexity of annotation and inter-annotator variability~\cite{yu2018learning,karimi2020deep,algan2021image,song2022learning}. This implies that some observed sample-label pairs may be incorrect, especially when dealing with large datasets. DCNNs can memorize randomly labeled samples~\cite{Zhang2017UnderstandingDL} and overfit noisy labels, leading to poor performance and generalizability. 
Therefore, it is crucial to improve the robustness of DCNNs against label noise.

Many research efforts have been devoted to learning with label noise, resulting in a number of solutions. Most methods attempt to filter out wrongly-labeled samples and then reduce their impact~\cite{han2018co,li2019dividemix,wang2022scalable,xia2020robust,kim2019nlnl,thulasidasan2019combating,patrini2017making,xia2022sample,Huang2022UncertaintyAwareLA}.
Although these methods avoid the complex modeling of label noise, they have limited reliability since the classifier learned on data with label noise, known as the noisy-label-trained classifier, may not be statistically consistent or, in other words, may not approach the optimal classifier defined on clean risk~\cite{li2021provably}.

To address this limitation, researchers are opting to explicitly model label noise~\cite{cheng2022instance,li2021provably}. This is achieved by converting the noisy posterior probability $P(Y_i^*\mid X_i)$ into a clean posterior probability $P(Y_i\mid X_i)$, also known as the \textit{Label Distribution (LD)}, using a transition matrix, where $Y_i^*$ and $Y_i$ represent the observed label and unseen ground truth label of a sample $\textit{\textbf{X}}_i$, respectively. The resulting clean posterior probability $P(Y_i\mid X_i)$ is then utilized to supervise the training process~\cite{hendrycks2018using,li2021provably,xia2019anchor,shu2020meta,yao2020dual,cheng2020learning,cheng2022instance,berthon2021confidence,xia2020part,yang2021estimating,kye2022learning}.
These methods are theoretically guaranteed to be statistically consistent.

Estimating the noise transition matrix accurately is the most essential and challenging step in these methods~\cite{xia2019anchor}, due to the ground truth of samples being inaccessible.
To achieve a reliable estimation, it is typically assumed that a small set of anchor points, $i.e.$, correctly labeled samples, are available~\cite{hendrycks2018using,shu2020meta,berthon2021confidence,xia2020part,kye2022learning}. 
However, this assumption greatly limits the application of these methods, since it is often difficult to determine which sample label is truly correct in a dataset with label noise.
To overcome this limitation, other methods utilize reliable data points extracted from the training set as pseudo anchor points~\cite{xia2019anchor,yao2020dual,cheng2022instance}. 
These methods depend heavily on the quality of pseudo anchor points, which is difficult to ensure in practice.
\textcolor{revise_text}{
To address these issues, we propose an alternative approach that leverages inter-class correlation to approximate the noise transition matrix, based on the observation that classes with greater feature similarity exhibit a higher propensity for mislabeling~\cite{wei2021learning}. 
For instance, within the CIFAR-10N dataset~\cite{wei2021learning}, which includes a clean-label set and five noisy-label sets, the mislabeling probability for similar classes (such as mistaking an automobile for a truck, a cat for a dog, or a deer for a horse) is markedly higher compared to dissimilar classes. 
Moreover, we have empirically validated this approximation by computing the Pearson and Spearman's Rank Correlation Coefficients between the class correlation matrix and the noise transition matrices obtained from CIFAR-10N. 
As shown in Fig.~\ref{fig: intro}, these coefficients, being significantly greater than 0.0 and nearing 1.0, indicate a strong positive linear and monotonic relationship between the two types of matrices. 
The higher Spearman's rank correlation coefficient also suggests that a high level of consistency in the rank order of data points between two matrices, supporting that inter-class similarity can effectively reflect the mislabeling probability.}

Moreover, label noise is dependent on the instance in real-world scenarios~\cite{wei2021learning}. For example, even if class $i$ is most similar to class $j$ among all classes, there may still be some samples in class $i$ whose closest match does not belong to class $j$.
Previous methods often estimate a global noise transition matrix for simplification~\cite{hendrycks2018using,xia2019anchor,shu2020meta,yao2020dual,li2021provably,cheng2022class}, and the global noise transition matrix fails to provide accurate label transition information for the samples in our previous example. 
To address this issue and approximate the instance-dependent noise transition matrix more accurately, we advocate calculating the inter-class correlation matrix at the instance level using a mini-batch of samples instead of all training samples.
\textcolor{revise_text}{This method avoids diluting local, sample-specific characteristics that might occur when using all training data.}

In this paper, we propose an Instance-dependent Label Distribution Estimation (ILDE) method for learning with label noise in image classification. 
ILDE models the inter-class correlation to approximate instance-dependent noise transition matrices.
Specifically, the Gram Matrix is employed to represent the inter-class correlation. 
Thus, we can approximate the instance-dependent noise transition matrix with the Gram Matrix calculated based on samples in a mini-batch, without using (pseudo) anchor points.
The noisy posterior probability of each sample is estimated under the supervision of noisy labels, and the instance-dependent LD is inferred based on this probability and the estimated noise transition matrix. To minimize the impact of noise, we sharpen the instance-dependent LD.
Recently, researchers have witnessed that DCNNs tend to fit clean and simple patterns before memorizing noisy labels~\cite{liu2020early}.
Therefore, we utilize the temporal ensembling technique~\cite{Laine2017TemporalEnsembling} that ensembles historical estimations of instance-dependent LD to improve supervision.

The contributions of this work are three-fold.

\begin{enumerate}
\item{We propose an ILDE method that estimates instance-dependent LD as supervision to enhance the robustness of DCNNs training against label noise.}
\item{Our ILDE approximates the instance-dependent noise transition matrices by modeling the inter-class correlation based on samples in a mini-batch, without relying on (pseudo) anchor points.}
\item{Experimental results on both synthetic and real-world noisy datasets demonstrate that our ILDE outperforms other existing methods that handle the label noise.}
\end{enumerate}

\section{Related Work}
\subsection{Learning with Label Noise}
There are two types of methods for learning with label noise: statistically inconsistent and statistically consistent.

Statistically inconsistent methods are designed to identify and mitigate the impact of noisy labels. 
These methods usually employ one of three techniques to identify such samples: the small loss trick~\cite{han2018co,shen2019learning,li2019dividemix,wang2022scalable}, which involves regarding samples with large loss value as noisy samples;
the prediction confidence trick~\cite{kim2019nlnl}, which relies on measuring the model's confidence in its predictions and treats samples with low confidence as noisy ones; 
or the early learning phenomenon trick~\cite{xia2020robust,liu2020early}, which identifies noisy labels by leveraging the early learning phenomena that DCNNs tend to fit clean and simple patterns before memorizing noisy labels~\cite{liu2020early}.
Once identified, there are several ways to reduce the impact of noisy samples, including discarding them, treating them as unlabeled samples for semi-supervised learning~\cite{nguyen2019self,jiang2018mentornet,wang2022scalable,xia2020robust}, 
assigning them pseudo labels~\cite{kim2019nlnl}, and adopting a sample weighting scheme~\cite{thulasidasan2019combating,patrini2017making,Wei2022ToSO}.
However, these methods have a major drawback - they cannot guarantee the reliability of classifiers trained on noisy labels. Additionally, it is challenging for these methods to differentiate between informative hard samples and harmful mislabeled ones~\cite{sukhbaatar2015training}.

In contrast, methods that are statistically consistent model label noise either implicitly or explicitly. Implicit modeling of label noise can be achieved by using a noise-tolerant loss function~\cite{ghosh2017robust,zhang2018generalized,wang2019symmetric,lyu2019curriculum,ma2020normalized,yao2020deep,zhou2021asymmetric} and regularization techniques~\cite{menon2019can,zhou2021learning}. 
For example, when the symmetric condition is met, risk minimization \textit{w.r.t.} the symmetric cross-entropy loss~\cite{wang2019symmetric} becomes tolerant to noise.
Explicit modeling of label noise aims to estimate the transition matrix for noisy labels and uses it and the noisy posterior probability to infer a clean posterior probability ~\cite{hendrycks2018using,yao2020dual,li2021provably,xia2019anchor,berthon2021confidence,xia2020part,cheng2022instance,cheng2020learning}.
Although these methods are theoretically guaranteed to be statistically consistent, they heavily rely on accurate estimation of the transition matrix. % for noisy labels.

\subsection{Estimation of Noise Transition Matrix}
The noise transition matrix connects the clean and noisy posterior probability. For binary classification tasks, Natarajan et al.~\cite{natarajan2013learning} suggest a cross-validation approach to estimate this matrix. In multi-classification tasks, researchers have learned the noise transition matrix by utilizing anchor points~\cite{patrini2017making,yu2018efficient,yao2020towards,hendrycks2018using,shu2020meta,berthon2021confidence,xia2020part,kye2022learning}.
If anchor points are not available, pseudo anchor points can be extracted from noisy training data based on estimated noisy class posterior~\cite{xia2019anchor}, estimated intermediate class posterior~\cite{yao2020dual}, or Bayes optimal labels~\cite{cheng2022instance,yang2021estimating}. These pseudo anchor points are then used to estimate the noise transition matrix.
In contrast to these methods that require identifying anchor or pseudo-anchor points, we propose estimating inter-class correlation as an approximation of the noise transition matrix, due to the observation that mislabeling probability is highly correlated with inter-class similarity. 
Our approach has only one training stage and does not require the identification of any anchors or pseudo-anchors.
\textcolor{revise_text}{Note that although VolMinNet~\cite{li2021provably} and our ILDE share the common approach of end-to-end training without anchor points, they diverge in their objectives: VolMinNet targets class-dependent noise transition matrices, which may not capture the heterogeneity of noise across instances within a class, while ILDE is tailored to approximate instance-dependent noise transition matrices, providing a more accurate representation of label noise that is mostly instance-dependent. The ability of ILDE to model instance-dependent noise gives it a distinct advantage over VolMinNet and other class-dependent methods, enabling superior performance in the presence of complex label noise patterns.}

\begin{figure*}[h]
\centering
\includegraphics[width=1.7\columnwidth]{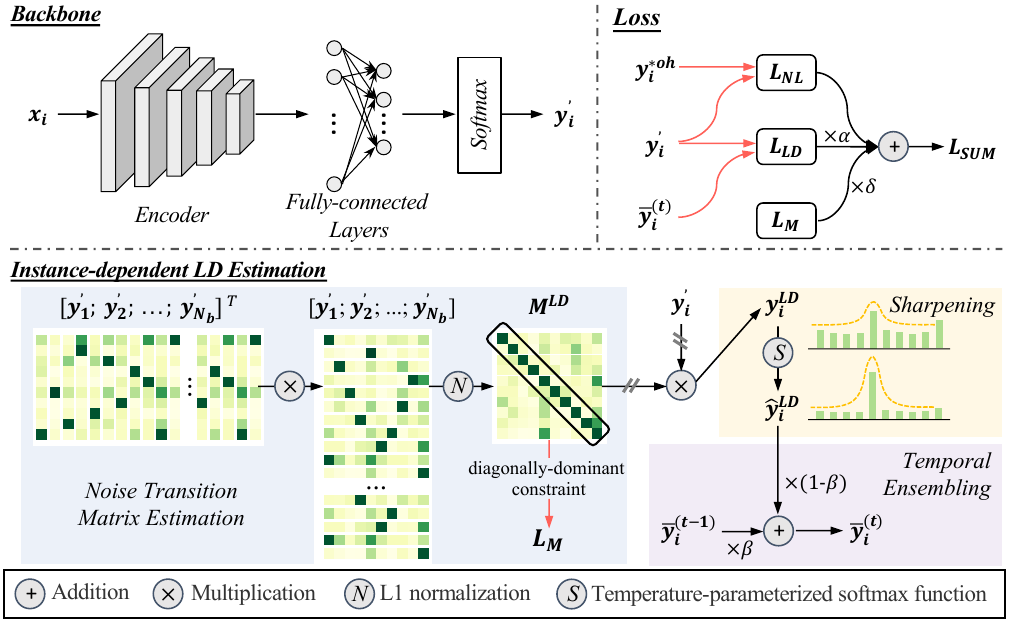}
\caption{
The overview of our Instance-dependent Label Distribution Estimation (ILDE) method.
The backbone is used for image classification.
The Instance-dependent LD Estimation Block is used for producing the instance-dependent LD for each input image, and the noise transition matrix estimation is based on all samples in a mini-batch ($N_b$ is the batch size).
The loss contains three terms, including the noisy supervision ($L_{NL}$), the instance-dependent LD supervision ($\mathcal{L}_{LD}$), and the diagonally-dominant constraint ($\mathcal{L}_M$).
For the input image $\textbf{\textit{x}}_i$, $\textbf{\textit{y}}_i^{'}$ is the output of the backbone.
$\textbf{\textit{y}}_i^{*oh}$ means one-hot observed label of $\textbf{\textit{x}}_i$.
$\textbf{\textit{y}}_i^{LD}$, $\hat{\textbf{\textit{y}}}_i^{LD}$ and $\bar{\textbf{\textit{y}}}_i^{(t)}$ are the estimated LD, the sharpened one and the ensembled one, respectively.
$\bar{\textbf{\textit{y}}}_i^{(t)}$ is calculated as the weighted summation of the historical target $\bar{\textbf{\textit{y}}}_i^{(t-1)}$ and the sharpened LD $\hat{\textbf{\textit{y}}}_i^{LD}$ at the $t$-th epoch.
}
\label{fig: overview}
\end{figure*}

\section{Preliminaries}

We consider a $K$-class image classification problem.
Let $\mathcal{X} \subset \mathbb{R} ^ {D}$ represent the image or feature space, where $D$ is the dimension of feature vectors.
Let $\mathcal{Y}=\left \{ 1,2,...,K \right \}$ represent the label space.
The noisy training dataset with $N$ image-label pairs is denoted by $\tilde{D} = \left \{ \left ( \textbf{\textit{x}}_i, y^{*}_i \right )  \right \} ^ {N} _{i=1}$, where $y^{*}_{i}$ is the observed label of $\textbf{\textit{x}}_i$, and $\left ( \textbf{\textit{x}}_i, y^{*}_i \right )$ is drawn $i.i.d.$ from a distribution over $\mathcal{X} \times \mathcal{Y}$.
We denote the unseen ground truth label of $\textbf{\textit{x}}_i$ as $y_i$. 
For the convenience of the mathematical formulation, we convert $y^{*}_i$ and $y_{i}$ into one-hot label vectors, denoted by $\textbf{\textit{y}}^{*oh}_{i}$ and $\textbf{\textit{y}}^{oh}_i$, respectively.
In this work, we aim to create a classifier for precise image classification via training on the noisy dataset $\tilde{D}$.

For each image $\textbf{\textit{x}}_i$, the probability that its observed label equals class $k$ is denoted as $p_i^k$, and its label distribution (\textit{i.e.}, the LD) is presented as $\left[ p_i^1, p_i^2,...,p_i^K \right]$.
The noise rate of image $\textbf{\textit{x}}_i$ is the probability that the observed label $y^{*}_i$ does not equal to the ground truth label $y_{i}$, calculated as

\begin{equation}
p_i= {\textstyle \sum_{k\ne y_i} p_i^k},
\end{equation}

Label noise can be roughly grouped into symmetric noise and asymmetric noise. For symmetric noise (\textit{a.k.a.}, uniform noise), the probabilities of a sample $\textbf{\textit{x}}_i$ being assigned to any wrong labels are equal, shown as follows
\begin{equation}
p_i^k \equiv \frac{p_i}{K-1}, k \ne y_i.
\end{equation}
For asymmetric noise, each incorrect label is assigned to $\textbf{\textit{x}}_i$ with a unique probability.
Moreover, if each sample within a class has the same LD, the noise is class-dependent, which oversimplifies the reality of label noise. 
In real practice, the distribution of label noise depends not only on the class label but also on the sample itself. It means that label noise is instance-dependent~\cite{wei2021learning}.
Therefore, each sample possesses its unique LD.
To combat the label noise using improved supervision, we estimate the instance-dependent LD for each sample.

\section{Methodology}
\label{sec: methodology}

In this section, we present the ILDE method and how to exploit the instance-dependent LD to train a robust classifier.
We adopt a backbone for image classification under the supervision of the observed label and the estimated instance-dependent LD.
The latter is produced by the Instance-dependent LD Estimation Block in three steps, $i.e.$, noise transition matrix estimation, sharpening, and temporal ensembling.
The overview of our ILDE method is shown in Fig.~\ref{fig: overview}.
We now delve into its details.

\subsection{Backbone}
The image classification backbone used for this study consists of an encoder $\mathcal{F}_E(\Theta_E)$, fully connected layers $\mathcal{F}_{FC}(\Theta_{FC})$, and a softmax function $\mathcal{S}$, where $\Theta_E$ and $\Theta_{FC}$ denote the parameters of encoder and fully connected layers respectively.
Given an input image $\textbf{\textit{x}}_i$, the soft prediction made by the backbone can be formally given as
\begin{equation}
    \textbf{\textit{y}}_i^{'}=\mathcal{S}(\mathcal{F}_{FC}(\mathcal{F}_E(\textbf{x}_i, \Theta_E), \Theta_{FC})),
\end{equation}
where $\textbf{\textit{y}}_i^{'} \in [0,1]^{K}$ can be treated as the estimated noisy posterior probabilities, which is supervised by the observed one-hot noisy label $\textbf{\textit{y}}_i^{*oh}$ (\textit{i.e.}, the noisy supervision) using the following cross-entropy loss
\begin{equation}
    \mathcal{L}_{NL}=- \textbf{\textit{y}}_{i}^{*oh}\log_{}{\textbf{\textit{y}}'_{i}}.
\label{eq: loss noisy label}
\end{equation}
Note that the architectures of the encoder and fully connected layers are tailored according to the training dataset.

\subsection{Instance-dependent LD Estimation Block}

Besides the observed noisy label, the image classification in our method is also supervised by the estimated instance-dependent LD, which is produced by the Instance-dependent LD Estimation Block in three steps.
First, we approximate the instance-dependent noise transition matrix based on the inter-class correlation and calculate the instance-dependent LD using the estimated noise transition matrix and the prediction given by the backbone.
Second, the estimated LD is sharpened by temperature-parameterized softmax to reduce the impact of noise.
Third, the sharpened LD is further enhanced by using the temporal ensembling technique.

\subsubsection{Noise Transition Matrix Estimation}
The noisy posterior probability $\textbf{\textit{y}}_i^{'}$ is estimated under the supervision of the noisy label $\textbf{\textit{y}}_i^{*oh}$ (see Eq.~\ref{eq: loss noisy label}).
Once we estimate the instance-dependent noise transition matrix, we can obtain the instance-dependent LD of each sample.
Since it has been recognized that a sample is more likely to be misassigned to the class that is similar to the ground truth one~\cite{wei2021learning,cohen2020separability,beyer2020we}, the mislabeling probability is highly correlated to the inter-class correlation.
Hence, we use the inter-class correlation to approximate the noise transition matrix. 
Specifically, the Gram Matrix, which has been widely used to characterize the spatial/channel/sample-wise correlation~\cite{liu2020semi}, is employed to represent the class-wise correlation.
To improve our accuracy, we calculate the Gram matrix using the soft predictions of the samples in a mini-batch instead of all training samples.
Let the soft predictions of the samples in a mini-batch be denoted by $\textbf{\textit{y}}' = [\textbf{\textit{y}}_1^{'};\textbf{\textit{y}}_2^{'};...;\textbf{\textit{y}}_{N_b}^{'}] \in [0,1]^{N_b \times K}$, where $N_b$ is the size of a mini-batch. The instance-dependent noise transition matrix $M^{LD} \in [0,1]^{K \times K}$ can be estimated as follows
\begin{equation}
    M^{LD} = \mathcal{N} ( \textbf{\textit{y}}'^T \times \textbf{\textit{y}}' )
\label{eq: noise transition matrix}
\end{equation}
where $\mathcal{N}$ represents the L1-normalization on each row of the column Gram matrix of the soft prediction $\textbf{\textit{y}}'$. Additionally, we introduce a regularization term $\mathcal{L}_M$ to make $M^{LD}$ is diagonally-dominant, shown as follows
\begin{equation}
    \mathcal{L}_M=\left \| M^{LD} \right \| _{2}^{2} - \left \| diag(M^{LD}) \right \| _{2}^{2}
\end{equation}
which represents the square of the L2 norm of all elements in $M^{LD}$ except the diagonal elements.

Based on the estimated noise transition matrix $M^{LD}$ and the estimated noisy posterior probability $\textbf{\textit{y}}_i^{'}$, the instance-dependent LD of $\textbf{\textit{x}}_i$ is calculated as
\begin{equation}
    \textbf{\textit{y}}_i^{LD}=\textbf{\textit{y}}_i^{'} \times M^{LD^T}.
\end{equation}
Note that we clip the gradient of $M^{LD}$ and $\textbf{\textit{y}}_i^{'}$ before calculating the matrix multiplication between them.

\subsubsection{Sharpening}
Intuitively, a sharp LD means less impact of noise on the label distribution, leading to more accurate image classification.
Therefore, we utilize the temperature-parameterized softmax function to sharpen the estimated instance-dependent LD $\textbf{\textit{y}}_i^{LD}$, shown as follows
\begin{equation}
    \hat{\textbf{\textit{y}}}^{LD}_{i,k}=\frac{\exp (\textbf{\textit{y}}^{LD}_{i,k}/\tau)}{\sum_{j}^{}  \exp (\textbf{\textit{y}}^{LD}_{i,j}/\tau )}
\label{eq: tau}
\end{equation}
where $\textbf{\textit{y}}^{LD}_{i,k}$ and $\hat{\textbf{\textit{y}}}^{LD}_{i,k}$ are the $k$-th element in the estimated instance-dependent LD before and after sharpening, and $\tau$ is a temperature factor.
When $0 < \tau < 1$, the input vector $\textbf{\textit{y}}_i^{LD}$ is sharpened; when $\tau > 1$, the input vector $\textbf{\textit{y}}_i^{LD}$ is smoothed. 

\subsubsection{Temporal Ensembling}
Meanwhile, to improve the robustness of our estimation, the sharpened instance-dependent LD $\hat{\textbf{\textit{y}}}_i^{LD}$ is further polished using the temporal ensembling technique~\cite{Laine2017TemporalEnsembling, yang2022survey}.
Let the running average of $\hat{\textbf{\textit{y}}}_i^{LD}$ in $t$-th epoch be denoted by $\bar{\textbf{\textit{y}}}_i^{(t)}$, which can be calculated as follows
\begin{equation}
    \bar{\textbf{\textit{y}}}_i^{(t)} = \beta \times \bar{\textbf{\textit{y}}}_i^{(t-1)} + (1- \beta) \times \hat{\textbf{\textit{y}}}_i^{LD}
\label{eq: beta}
\end{equation}
where $0 \le \beta < 1$ is  the momentum. Note that $\bar{\textbf{\textit{y}}}_i^{(0)}$ is an all-zero tensor.

\noindent \textbf{LD Supervision.} 
Since the observed label $y^{*}_i$ is noisy, using it to supervise image classification (see Eq.~\ref{eq: loss noisy label}) may provide incorrect supervision when $y^{*}_i$ is not true. 
To address this issue, we also use the estimated instance-dependent LD $\bar{\textbf{\textit{y}}}_i^{(t)}$ as the supervision. 
Specifically, we introduce the loss $\mathcal{L}_{LD}$ to maximizes the inner product between the prediction $\textbf{\textit{y}}_i^{'}$ and $\bar{\textbf{\textit{y}}}_i^{(t)}$, shown as follows
\begin{equation}
    \mathcal{L}_{LD}=-\log_{}{(1-\bar{\textbf{\textit{y}}}_{i}^{(t)}\textbf{\textit{y}}'_{i})} 
\end{equation}
Note that, although the cross-entropy loss and Kullback-Leibler divergence can be utilized to push $\textbf{\textit{y}}_i^{'}$ close to $\bar{\textbf{\textit{y}}}_i^{(t)}$ as well, the inner product is more effective and performs better~\cite{liu2020early}.

\subsection{Loss Function}
The total loss of our ILDE method can be calculated as
\begin{equation}
    \mathcal{L}_{SUM}=\mathcal{L}_{NL} + \alpha \times \mathcal{L}_{LD} + \delta \times \mathcal{L}_M,
\label{eq: alpha_delta}
\end{equation}
where $\alpha$ and $\delta$ are weighting coefficients.

\section{Experiments}
\subsection{Datasets}
\subsubsection{Synthetic Noisy Datasets}
Two synthetic noisy datasets were constructed based on MNIST~\cite{lecun1998gradient} and CIFAR-10~\cite{krizhevsky2009learning} for this study.
The MNIST\footnote{\url{http://yann.lecun.com/exdb/mnist/}} dataset contains 60,000 training images and 10,000 testing images, and CIFAR-10\footnote{\url{https://www.cs.toronto.edu/~kriz/cifar.html}} consists of 50,000 training images and 10,000 test images.
We considered two types of noisy labels: (1) symmetric noise, where the true label is randomly altered to any other class label with an equal probability; and (2) asymmetric noise, where the true label is more likely to be corrupted to similar classes. Both types of synthetic noise are class-dependent.

\subsubsection{Real-world Noisy Datasets}
\textcolor{revise_text}{
Three real-world noisy datasets, CIFAR-10N, CIFAR-100N~\cite{wei2021learning} and Clothing1M~\cite{xiao2015learning} were utilized in this study.
CIFAR-10N and CIFAR-100N, collectively referred to as CIFAR-N\footnote{\url{http://noisylabels.com/}}, are established by equipping the training set of CIFAR-10 and CIFAR-100 with human-annotated noisy labels collected from Amazon Mechanical Turk while keeping the test set clean.}
Each dataset consists of 50,000 training images and 10,000 test images.
CIFAR-10N includes five sets of noisy labels for training data, Aggregate (noise rate=$9.03\%$), Random 1 (noise rate=$17.23\%$), Random 2 (noise rate=$18.12\%$), Random 3 (noise rate=$17.64\%$), and Worst (noise rate=$40.21\%$). CIFAR-100N has only one set of noisy labels for training data with a noise rate of $40.20\%$.
It is worth noting that previous research has shown that the human noise in these datasets is instance-dependent~\cite{wei2021learning}.
\textcolor{revise_text}{The Clothing1M dataset comprises one million images with real-world noisy labels. Additionally, there are three distinct sets of 50k, 14k, and 10k images with clean labels for training, validation, and testing purposes. 
For this study, we focused on the noisy set, which has an estimated noise rate of 38.5\%, using it exclusively for training.}

\subsection{Implementation Details}
\textcolor{revise_text}{In our ILDE method, the backbone is tailored according to the dataset.
For MNIST, the backbone consists of two convolutional layers and two FC layers.
For CIFAR-10 and CIFAR-10N, the backbone consists of six convolutional layers and two FC layers.
For CIFAR-100N, we use ResNet-34~\cite{he2016deep} as the backbone.
For Clothing1M, we adopt a ResNet50 network pre-trained on the ImageNet dataset as the backbone.
The SGD optimizer with a momentum of 0.9 was used for all experiments.
The initial learning rate is set to 0.01 for MNIST, CIFAR-10 and CIFAR-10N, set to 0.1 for CIFAR-100N, and set to 0.001 and then divided by 10 after 5th epoch for Clothing1M.
The batch size was set to 32 for Clothing1M and set to 128 for other datasets.
The weight decay was set to 1e-3, 1e-4, 1e-4, 1e-5, 0.005 for MNIST, CIFAR-10, CIFAR-10N, CIFAR-100N and Clothing1M, respectively.
The model was trained for 50 epochs on MNIST, 120 epochs on CIFAR-10 and CIFAR-10N, 200 epochs on CIFAR-100N, 20 epochs on Clothing1M.
Random cropping and random horizontal flipping were adopted to increase the data diversity of CIFAR10, CIFAR-10N, and CIFAR-100N.
The experiments were performed on the PyTorch framework using a workstation with two NVIDIA RTX 3080Ti GPUs. 
Results over five trials on all datasets except Clothing1M, for which the result is over one trial, are reported.}

\begin{table*}[t]
\begin{center}
\caption{Test accuracy (\%, mean$\pm$standard deviation) of our ILDE, baseline, and competing methods on MNIST and CIFAR-10 with clean labels or symmetric label noise ($noise\ rate \in [0.2, 0.4, 0.6, 0.8]$). The best and second best results in each column are highlighted in \textbf{bold} and \underline{underline}, respectively.}
\label{tab: comparison_sym}
\small
\setlength\tabcolsep{5.0pt}
\begin{tabular}{ccccccc}
\hline \hline
\multirow{2}{*}{Dataset}  & \multicolumn{1}{c}{\multirow{2}{*}{Method}} & \multirow{2}{*}{Clean} & \multicolumn{4}{c}{Symmetric Noise Rate}                                                                                                                    \\ \cline{4-7} 
                          & \multicolumn{1}{c}{}                        &                        & \multicolumn{1}{c}{0.2}                   & \multicolumn{1}{c}{0.4}                   & \multicolumn{1}{c}{0.6}                   & 0.8                   \\ \hline
\multirow{8}{*}{MNIST}    & Baseline                                           & 99.08 $\pm $ 0.09           & \multicolumn{1}{c}{86.06 $\pm $ 0.76}          & \multicolumn{1}{c}{66.70 $\pm $ 1.25}          & \multicolumn{1}{c}{44.40 $\pm $ 1.66}          & 20.89 $\pm $ 1.17          \\  
                          & SCE                                     & 99.01 $\pm $ 0.25           & \multicolumn{1}{c}{88.76 $\pm $ 0.49}          & \multicolumn{1}{c}{69.04 $\pm $ 1.42}          & \multicolumn{1}{c}{47.29 $\pm $ 0.62}          & 21.19 $\pm $ 1.02          \\  
                          & ELR                                     & 99.18 $\pm $ 0.06           & \multicolumn{1}{c}{99.05 $\pm $ 0.06} & \multicolumn{1}{c}{98.87 $\pm $ 0.04}          & \multicolumn{1}{c}{98.54 $\pm $ 0.12}          & 91.37 $\pm $ 1.01          \\  
                          & AUL                                     & 99.18 $\pm $ 0.09           & \multicolumn{1}{c}{99.02 $\pm $ 0.05}          & \multicolumn{1}{c}{98.88 $\pm $ 0.11}          & \multicolumn{1}{c}{98.14 $\pm $ 0.20}          & 97.14 $\pm $ 0.02          \\  
                          & AGCE                                    & 99.19 $\pm $ 0.07           & \multicolumn{1}{c}{99.05 $\pm $ 0.06} & \multicolumn{1}{c}{98.92 $\pm $ 0.07}          & \multicolumn{1}{c}{98.36 $\pm $ 0.09}          & 96.74 $\pm $ 0.15          \\  
                          & SR                                      & \underline{99.30 $\pm $ 0.06}  & \multicolumn{1}{c}{98.99 $\pm $ 0.10}          & \multicolumn{1}{c}{98.93 $\pm $ 0.08} & \multicolumn{1}{c}{98.69 $\pm $ 0.12} & 98.37 $\pm $ 0.19 \\  
                          & GLS                                      & \textbf{99.33 $\pm $ 0.06}  & \multicolumn{1}{c}{98.66 $\pm $ 0.06}          & \multicolumn{1}{c}{97.94 $\pm $ 0.15} & \multicolumn{1}{c}{95.92 $\pm $ 0.31} & 76.65 $\pm $ 1.59  \\  
                          
                          & VolMinNet                                      & 98.99 $\pm$ 0.13  & \multicolumn{1}{c}{98.87 $\pm$ 0.08}          & \multicolumn{1}{c}{98.74 $\pm$ 0.09} & \multicolumn{1}{c}{98.46 $\pm$ 0.22} & 98.17 $\pm$ 0.19  \\  
                          
                          & MEIDTM                                      & 99.04 $\pm $ 0.64  & \multicolumn{1}{c}{99.05 $\pm $ 0.37}          & \multicolumn{1}{c}{98.95 $\pm $ 0.24} & \multicolumn{1}{c}{98.70 $\pm $ 0.32} & 98.35 $\pm $ 0.39  \\  
                          & Ours                                         & 99.18 $\pm $ 0.08           & \multicolumn{1}{c}{\underline{99.06 $\pm $ 0.09}} & \multicolumn{1}{c}{\underline{98.95 $\pm $ 0.07}} & \multicolumn{1}{c}{\underline{98.81 $\pm $ 0.07}} & \underline{98.39 $\pm $ 0.18} \\ 
                          & Ours$^*$                                      & 99.24 $\pm $ 0.07           & \multicolumn{1}{c}{\textbf{99.14 $\pm $ 0.04}} & \multicolumn{1}{c}{\textbf{99.04 $\pm $ 0.04}} & \multicolumn{1}{c}{\textbf{98.84 $\pm $ 0.09}} & \textbf{98.57 $\pm $ 0.04} \\ \hline
\multirow{8}{*}{CIFAR-10} & Baseline                                           & 90.12 $\pm $ 0.17           & \multicolumn{1}{c}{71.87 $\pm $ 0.55}          & \multicolumn{1}{c}{54.29 $\pm $ 1.23}          & \multicolumn{1}{c}{36.43 $\pm $ 1.18}          & 17.62 $\pm $ 0.32          \\  
                          & SCE                                     & 90.63 $\pm $ 0.26           & \multicolumn{1}{c}{77.01 $\pm $ 0.45}          & \multicolumn{1}{c}{59.16 $\pm $ 0.97}          & \multicolumn{1}{c}{40.23 $\pm $ 0.41}          & 17.91 $\pm $ 0.32          \\  
                          & ELR                                     & \textbf{91.18 $\pm $ 0.28}  & \multicolumn{1}{c}{88.68 $\pm $ 0.17} & \multicolumn{1}{c}{85.90 $\pm $ 0.17} & \multicolumn{1}{c}{77.04 $\pm $ 0.35}          & 35.75 $\pm $ 1.42          \\  
                          & NCE+AUL                                 & 90.06 $\pm $ 0.11           & \multicolumn{1}{c}{88.06 $\pm $ 0.28}          & \multicolumn{1}{c}{84.97 $\pm $ 0.21}          & \multicolumn{1}{c}{77.95 $\pm $ 0.32}          & 55.54 $\pm $ 1.59 \\  
                          & NCE+AGCE                                & 89.82 $\pm $ 0.20           & \multicolumn{1}{c}{88.06 $\pm $ 0.20}          & \multicolumn{1}{c}{84.67 $\pm $ 0.66}          & \multicolumn{1}{c}{78.27 $\pm $ 0.77} & 47.55 $\pm $ 4.10          \\  
                          & SR                                      & 90.01 $\pm $ 0.06           & \multicolumn{1}{c}{88.04 $\pm $ 0.18}          & \multicolumn{1}{c}{84.81 $\pm $ 0.16}          & \multicolumn{1}{c}{78.02 $\pm $ 0.44}          & 50.61 $\pm $ 0.93          \\  
                          & GLS                                      & 90.10 $\pm $ 0.36           & \multicolumn{1}{c}{81.99 $\pm$ 0.64}          & \multicolumn{1}{c}{72.53 $\pm$ 0.71}          & \multicolumn{1}{c}{61.80 $\pm$ 0.64}          & 36.94 $\pm$ 1.88           \\  
                          & VolMinNet                                      & 89.92 $\pm$ 0.17  & \multicolumn{1}{c}{88.85 $\pm$ 0.29}          & \multicolumn{1}{c}{85.99 $\pm$ 0.52} & \multicolumn{1}{c}{78.34 $\pm$ 0.56} & 54.37 $\pm$ 0.48  \\  
                          & MEIDTM                                      & 90.04 $\pm $ 0.46  & \multicolumn{1}{c}{89.01 $\pm $ 0.37}          & \multicolumn{1}{c}{86.22 $\pm $ 0.68} & \multicolumn{1}{c}{78.51 $\pm $ 0.79} & 55.84 $\pm $ 0.95  \\  
                          & Ours                                         & 90.36 $\pm $ 0.16           & \multicolumn{1}{c}{\underline{89.35 $\pm $ 0.14}} & \multicolumn{1}{c}{\underline{87.03 $\pm $ 0.41}} & \multicolumn{1}{c}{\underline{79.34 $\pm $ 0.54}} & \underline{56.25 $\pm $ 0.82}          \\  
                          & Ours$^*$                                      & \underline{90.92 $\pm $ 0.25}           & \multicolumn{1}{c}{\textbf{89.53 $\pm $ 0.12}} & \multicolumn{1}{c}{\textbf{87.27 $\pm $ 0.46}} & \multicolumn{1}{c}{\textbf{81.35 $\pm $ 0.43}} & \textbf{59.10 $\pm $ 0.84}          \\ \hline \hline
\end{tabular}
\end{center}
\end{table*}

\begin{table*}[t]
\begin{center}
\caption{Test accuracy (\%, mean$\pm$standard deviation) of our ILDE, baseline, and competing methods on the MNIST dataset and CIFAR-10 dataset with asymmetric label noise ($noise\ rate \in [0.1, 0.2, 0.3, 0.4, 0.5] $). The best and second best results in each column are highlighted in \textbf{bold} and \underline{underline}, respectively.}
\label{tab: comparison_asym}
\small
\setlength\tabcolsep{5.0pt}
\begin{tabular}{clccccc}
\hline \hline
\multirow{2}{*}{Dataset}  & \multicolumn{1}{c}{\multirow{2}{*}{Method}} & \multicolumn{5}{c}{Asymmetric Noise Rate}                                                                                                                                                                \\ \cline{3-7} 
                          & \multicolumn{1}{c}{}                        & \multicolumn{1}{c}{0.1}                   & \multicolumn{1}{c}{0.2}                   & \multicolumn{1}{c}{0.3}                   & \multicolumn{1}{c}{0.4}                   & 0.5                   \\ \hline
\multirow{8}{*}{MNIST}    & Baseline                                           & \multicolumn{1}{c}{93.59 $\pm $ 0.59}          & \multicolumn{1}{c}{86.74 $\pm $ 1.06}          & \multicolumn{1}{c}{76.98 $\pm $ 0.99}          & \multicolumn{1}{c}{66.28 $\pm $ 1.42}          & 54.93 $\pm $ 1.88          \\  
                          & SCE                                     & \multicolumn{1}{c}{94.46 $\pm $ 0.29}          & \multicolumn{1}{c}{88.50 $\pm $ 1.03}          & \multicolumn{1}{c}{78.40 $\pm $ 1.28}          & \multicolumn{1}{c}{69.19 $\pm $ 2.27}          & 57.21 $\pm $ 1.49          \\  
                          & ELR                                     & \multicolumn{1}{c}{\underline{99.14 $\pm $ 0.07}} & \multicolumn{1}{c}{99.00 $\pm $ 0.11}          & \multicolumn{1}{c}{98.96 $\pm $ 0.09}          & \multicolumn{1}{c}{98.79 $\pm $ 0.09}          & 98.65 $\pm $ 0.10          \\  
                          & AUL                                     & \multicolumn{1}{c}{99.06 $\pm $ 0.03}          & \multicolumn{1}{c}{99.04 $\pm $ 0.07}          & \multicolumn{1}{c}{98.85 $\pm $ 0.16}          & \multicolumn{1}{c}{98.87 $\pm $ 0.11}          & 96.68 $\pm $ 3.78          \\  
                          & AGCE                                    & \multicolumn{1}{c}{99.10 $\pm $ 0.04}          & \multicolumn{1}{c}{99.07 $\pm $ 0.04} & \multicolumn{1}{c}{98.87 $\pm $ 0.10}          & \multicolumn{1}{c}{98.54 $\pm $ 0.22}          & 97.57 $\pm $ 0.45          \\  
                          & SR                                      & \multicolumn{1}{c}{98.79 $\pm $ 0.05}          & \multicolumn{1}{c}{98.99 $\pm $ 0.19}          & \multicolumn{1}{c}{99.03 $\pm $ 0.06} & \multicolumn{1}{c}{98.96 $\pm $ 0.08} & 98.85 $\pm $ 0.10 \\  
                          & GLS                                      & \multicolumn{1}{c}{98.94 $\pm $ 0.06}          & \multicolumn{1}{c}{98.71 $\pm $ 0.10}          & \multicolumn{1}{c}{98.38 $\pm $ 0.09} & \multicolumn{1}{c}{97.56 $\pm$ 0.10} & 95.92 $\pm$ 0.65  \\  
                          
                          & VolMinNet                                      & 99.03 $\pm$ 0.09  & \multicolumn{1}{c}{99.00 $\pm$ 0.15}          & \multicolumn{1}{c}{98.95 $\pm$ 0.07} & \multicolumn{1}{c}{98.81 $\pm$ 0.07} & 98.77 $\pm$ 0.19   \\
                          
                          & MEIDTM                                      & 99.07 $\pm $ 0.32  & \multicolumn{1}{c}{99.04 $\pm $ 0.26}          & \multicolumn{1}{c}{99.06 $\pm $ 0.12} & \multicolumn{1}{c}{98.85 $\pm $ 0.14} & 98.83 $\pm $ 0.21   \\  
                          & Ours                                         & \multicolumn{1}{c}{99.09 $\pm $ 0.07}          & \multicolumn{1}{c}{\underline{99.09 $\pm $ 0.06}} & \multicolumn{1}{c}{\underline{99.08 $\pm $ 0.06}} & \multicolumn{1}{c}{\underline{98.98 $\pm $ 0.06}} & \underline{98.97 $\pm $ 0.06} \\ 
                          & Ours$^*$                                      & \multicolumn{1}{c}{\textbf{99.20 $\pm $ 0.06}}          & \multicolumn{1}{c}{\textbf{99.16 $\pm $ 0.08}} & \multicolumn{1}{c}{\textbf{99.10 $\pm $ 0.05}} & \multicolumn{1}{c}{\textbf{99.00 $\pm $ 0.08}} & \textbf{98.97 $\pm $ 0.06} \\ \hline
\multirow{8}{*}{CIFAR-10} & Baseline                                           & \multicolumn{1}{c}{80.97 $\pm $ 0.39}          & \multicolumn{1}{c}{72.60 $\pm $ 0.98}          & \multicolumn{1}{c}{63.24 $\pm $ 0.74}          & \multicolumn{1}{c}{55.86 $\pm $ 0.82}          & 46.81 $\pm $ 0.57          \\  
                          & SCE                                     & \multicolumn{1}{c}{82.89 $\pm $ 0.40}          & \multicolumn{1}{c}{76.20 $\pm $ 0.51}          & \multicolumn{1}{c}{67.82 $\pm $ 0.82}          & \multicolumn{1}{c}{58.34 $\pm $ 0.76}          & 49.60 $\pm $ 1.01          \\  
                          & ELR                                     & \multicolumn{1}{c}{89.11 $\pm $ 0.22} & \multicolumn{1}{c}{88.43 $\pm $ 0.26} & \multicolumn{1}{c}{87.17 $\pm $ 0.14} & \multicolumn{1}{c}{84.31 $\pm $ 0.64} & 75.00 $\pm $ 0.32          \\  
                          & NCE+AUL                                 & \multicolumn{1}{c}{88.76 $\pm $ 0.28}          & \multicolumn{1}{c}{86.80 $\pm $ 0.57}          & \multicolumn{1}{c}{85.38 $\pm $ 0.38}          & \multicolumn{1}{c}{81.43 $\pm $ 0.60}          & 74.68 $\pm $ 1.17          \\  
                          & NCE+AGCE                                & \multicolumn{1}{c}{88.59 $\pm $ 0.23}          & \multicolumn{1}{c}{86.97 $\pm $ 0.55}          & \multicolumn{1}{c}{85.19 $\pm $ 0.35}          & \multicolumn{1}{c}{81.03 $\pm $ 1.09}          & 71.85 $\pm $ 0.58          \\  
                          & SR                                      & \multicolumn{1}{c}{88.89 $\pm $ 0.23}          & \multicolumn{1}{c}{87.16 $\pm $ 0.15}          & \multicolumn{1}{c}{85.25 $\pm $ 0.16}          & \multicolumn{1}{c}{81.97 $\pm $ 0.18}          & 76.58 $\pm $ 0.22 \\  
                          & GLS                                      & \multicolumn{1}{c}{84.60 $\pm$ 0.69}          & \multicolumn{1}{c}{80.27 $\pm$ 0.63}          & \multicolumn{1}{c}{76.40 $\pm$ 0.33}          & \multicolumn{1}{c}{71.53 $\pm$ 0.87}          & 65.07 $\pm$ 1.18  \\  
                          & VolMinNet                                      & 89.02 $\pm$ 0.41  & \multicolumn{1}{c}{87.56 $\pm$ 0.38}          & \multicolumn{1}{c}{85.97 $\pm$ 0.46} & \multicolumn{1}{c}{83.87 $\pm$ 0.34} & 75.79 $\pm$ 0.63  \\ 
                          & MEIDTM                                      & 89.18 $\pm $ 0.33  & \multicolumn{1}{c}{88.43 $\pm $ 0.27}          & \multicolumn{1}{c}{87.26 $\pm $ 0.31} & \multicolumn{1}{c}{84.50 $\pm $ 0.40} & 77.23 $\pm $ 0.57  \\  
                          & Ours                                         & \multicolumn{1}{c}{\underline{89.83 $\pm $ 0.19}} & \multicolumn{1}{c}{\underline{88.84 $\pm $ 0.30}} & \multicolumn{1}{c}{\underline{88.03 $\pm $ 0.24}} & \multicolumn{1}{c}{\underline{85.28 $\pm $ 0.46}} & \underline{77.88 $\pm $ 0.89} \\  
                          & Ours$^*$                                      & \multicolumn{1}{c}{\textbf{89.94 $\pm $ 0.18}} & \multicolumn{1}{c}{\textbf{89.20 $\pm $ 0.16}} & \multicolumn{1}{c}{\textbf{88.30 $\pm $ 0.23}} & \multicolumn{1}{c}{\textbf{85.89 $\pm $ 0.21}} & \textbf{80.74 $\pm $ 0.51} \\ \hline \hline
\end{tabular}
\end{center}
\end{table*}

\subsection{Comparisons with State-of-the-Art Methods}

\textcolor{revise_text}{We compared our ILDE with the baseline method and eight competing methods on two synthetic noisy datasets and three real-world noisy datasets.}
The baseline method is the classification backbone trained on the noisy dataset by minimizing the standard cross-entropy loss.
The competing methods include 
(1) the symmetric cross entropy (SCE) method~\cite{wang2019symmetric}, which combines the cross-entropy loss with a reverse cross-entropy term;
(2) the early-learning regularization (ELR) method~\cite{liu2020early}, which utilizes the ensemble of historical predictions as supervision;
(3) the methods using asymmetric loss functions (AUL, AGCE, NCE+AUL, and NCE+AGCE)~\cite{zhou2021asymmetric}, which makes the optimization direction shift to the loss term of class with the maximum weight;
(4) sparse regularization (SR) method~\cite{zhou2021learning}, which makes any loss be noise-robustness by restricting the prediction to the set of permutations over a fixed vector;
(5) generalized label smoothing (GLS) method~\cite{Wei2022ToSO}, which uses the positively or negatively weighted average of both the hard observed labels and uniformly distributed soft labels as the target labels;
(6) volume minimization network (VolMinNet)~\cite{li2021provably}, which optimizes the trainable noise transition matrix by minimizing the volume of the simplex formed by the columns of the noise transition matrix; 
(7) manifold embedding instance-dependent transition matrix (MEIDTM) method~\cite{cheng2022instance}, which proposes the practical assumption on the geometry of transition matrix to reduce the degree of freedom of transition matrix and make it stably estimable in practice; 
and 
All competing methods were re-implemented using their released codes and the reported optimal hyper-parameters.
The learning rate, batch size, weight decay, optimizer, epochs number, and backbone are kept the same as those in our ILDE for a fair comparison.

\subsubsection{Results on Synthetic Noisy Datasets}
The results of our ILDE methods and other competing methods on MNIST and CIFAR-10 were listed in Table~\ref{tab: comparison_sym} (with symmetric noise) and Table~\ref{tab: comparison_asym} (with asymmetric noise). 
The noise rate of symmetric noise ranges from 0.2 to 0.8, and the noise rate of asymmetric noise ranges from 0.1 to 0.5.
In the GLS method, the hyper-parameter is sensitive to noise rate, and hence is adjusted empirically for each noise rate.
In those competing methods, the hyper-parameter keeps unchanged for different noise rates.
Accordingly, our ILDE method was evaluated under these two settings, and we provide `Ours' and `Ours$^*$' in Table~\ref{tab: comparison_sym} and Table~\ref{tab: comparison_asym}.
In these tables, `Ours' represents the ILDE method with the same hyper-parameter setting for different noise rates, while `Ours$^*$' represents the ILDE methods with a unique hyper-parameter setting for each noise rate.

Although GLS and ELR perform better on the clean set of MNIST (GLS: 99.33\% v.s. Ours$^*$: 99.24\%) and CIFAR-10 (ELR: 91.18\% v.s. Ours$^*$: 90.92\%),
our ILDE achieves the best result on all symmetric noisy cases and most asymmetric noisy cases.
Moreover, our ILDE outperforms all competing methods in the most difficult situations by a clear margin.
On CIFAR-10 with the most severe symmetric noise (noise rate=0.8), ILDE outperforms the best competitor MEIDTM by 3.26\% (59.10\% v.s. 55.84\% and p-value=4.54e-4$<$0.05).
On CIFAR-10 with the most severe asymmetric noise (noise rate=0.5), our ILDE outperforms the best competitor MEIDTM by 3.51\% (80.74\% v.s. 77.23\% and p-value=7.65e-6$<$0.05).
In addition, it should be noted that the performance gain of our ILDE method on symmetric/asymmetric noisy MNIST is only 0.2\%/0.12\% (p-value=0.077/0.057). We believe it can be attributed to the fact that our performance is saturated on MNIST, since the average accuracy of our ILDE reaches 98.57\%/98.97\%. 

\begin{table*}[t]
\begin{center}
\caption{Test accuracy (\%, mean$\pm$standard deviation) of our ILDE, baseline, and competing methods on CIFAR-10N and CIFAR-100N. The best and second best results in each column are highlighted in \textbf{bold} and \underline{underline}, respectively.}
\label{tab: comparison_real_noise}
\small
\setlength\tabcolsep{3.5pt}
\begin{tabular}{lcccccc}
\hline \hline
\multicolumn{1}{c}{\multirow{2}{*}{Method}} & \multicolumn{5}{c}{CIFAR-10N}                                                                                                                               & CIFAR-100N   \\ \cline{2-6} \cline{7-7}
\multicolumn{1}{c}{}  & \multicolumn{1}{c}{Aggregate}    & \multicolumn{1}{c}{Random 1}     & \multicolumn{1}{c}{Random 2}     & \multicolumn{1}{c}{Random 3}     & Worst        & Noisy        \\ \hline
Baseline              & \multicolumn{1}{c}{84.05 $\pm$ 0.20} & \multicolumn{1}{c}{77.15 $\pm$ 0.76} & \multicolumn{1}{c}{76.07 $\pm$ 0.73} & \multicolumn{1}{c}{76.96 $\pm$ 0.65} & 59.10 $\pm$ 0.25 & 44.49 $\pm$ 1.40 \\ 
SCE                   & \multicolumn{1}{c}{84.98 $\pm$ 0.29} & \multicolumn{1}{c}{78.96 $\pm$ 0.44} & \multicolumn{1}{c}{78.59 $\pm$ 0.59} & \multicolumn{1}{c}{79.02 $\pm$ 0.48} & 62.60 $\pm$ 0.86 & 45.42 $\pm$ 1.49 \\ 
ELR                   & \multicolumn{1}{c}{88.53 $\pm$ 0.47} & \multicolumn{1}{c}{87.35 $\pm$ 0.27} & \multicolumn{1}{c}{87.49 $\pm$ 0.49} & \multicolumn{1}{c}{87.12 $\pm$ 0.19} & 81.19 $\pm$ 0.61 & 45.48 $\pm$ 0.59 \\ 
NCE+AUL               & \multicolumn{1}{c}{88.13 $\pm$ 0.43} & \multicolumn{1}{c}{86.28 $\pm$ 0.50} & \multicolumn{1}{c}{86.28 $\pm$ 0.68} & \multicolumn{1}{c}{86.54 $\pm$ 0.29} & 78.82 $\pm$ 0.73 & 52.01 $\pm$ 0.83 \\ 
NCE+AGCE              & \multicolumn{1}{c}{87.84 $\pm$ 0.16} & \multicolumn{1}{c}{86.51 $\pm$ 0.32} & \multicolumn{1}{c}{86.61 $\pm$ 0.25} & \multicolumn{1}{c}{86.20 $\pm$ 0.52} & 78.56 $\pm$ 0.89 & 54.82 $\pm$ 0.48 \\ 
SR                    & \multicolumn{1}{c}{88.38 $\pm$ 0.23} & \multicolumn{1}{c}{87.18 $\pm$ 0.18} & \multicolumn{1}{c}{86.95 $\pm$ 0.20} & \multicolumn{1}{c}{87.04 $\pm$ 0.18} & 79.51 $\pm$ 0.34 & 52.45 $\pm$ 0.30 \\ 
GLS                   & \multicolumn{1}{c}{84.93 $\pm$ 0.44} & \multicolumn{1}{c}{81.93 $\pm$ 0.45} & \multicolumn{1}{c}{81.83 $\pm$ 0.53} & \multicolumn{1}{c}{81.77 $\pm$ 0.48} & 72.86 $\pm$ 1.12 & 46.97 $\pm$ 0.94 \\ 
VolMinNet             & \multicolumn{1}{c}{88.83 $\pm$ 0.22} & \multicolumn{1}{c}{87.30 $\pm$ 0.14} & \multicolumn{1}{c}{87.14 $\pm$ 0.17} & \multicolumn{1}{c}{87.01 $\pm$ 0.25} & 80.07 $\pm$ 0.47 & 54.69 $\pm$ 0.46 \\ 
MEIDTM                & \multicolumn{1}{c}{89.10 $\pm$ 0.24} & \multicolumn{1}{c}{87.89 $\pm$ 0.12} & \multicolumn{1}{c}{87.93 $\pm$ 0.17} & \multicolumn{1}{c}{87.54 $\pm$ 0.21} & 81.43 $\pm$ 0.30 & 55.06 $\pm$ 0.41 \\ 
Ours                  & \multicolumn{1}{c}{\underline{89.16 $\pm$ 0.15}} & \multicolumn{1}{c}{\underline{88.52 $\pm$ 0.60}} & \multicolumn{1}{c}{\underline{88.52 $\pm$ 0.15}} & \multicolumn{1}{c}{\underline{88.19 $\pm$ 0.57}} & \underline{82.74 $\pm$ 0.69} & \underline{55.95 $\pm$ 0.47} \\ 
Ours*                 & \multicolumn{1}{c}{\textbf{89.49 $\pm$ 0.16}} & \multicolumn{1}{c}{\textbf{88.52 $\pm$ 0.60}} & \multicolumn{1}{c}{\textbf{88.55 $\pm$ 0.19}} & \multicolumn{1}{c}{\textbf{88.52 $\pm$ 0.15}} & \textbf{82.74 $\pm$ 0.69} & \textbf{57.83 + 0.61} \\ \hline \hline
\end{tabular}
\end{center}
\end{table*}

\begin{table}[t]
\centering
% \color{blue}
\setlength\tabcolsep{20.0pt}
\caption{
\textcolor{revise_text}{Test accuracy (\%) of our ILDE, baseline and competing methods on the Clothing1M dataset. The best and second-best results are highlighted in \textbf{bold} and \underline{underline}, respectively.}
}
\label{tab: comparison_Clothing1M}
\begin{tabular}{l|c}
\hline \hline
Method   & Accuracy \\ \hline \hline
Baseline & 68.29    \\ \hline
ELR      & 68.65    \\ \hline
SR       & 69.15    \\ \hline
MEIDTM   & 70.62    \\ \hline
TNLPAD   & 71.51    \\ \hline
Ours     & 72.45    \\ \hline \hline
\end{tabular}
\end{table}

\subsubsection{Results on Real-world Noisy Datasets}
\textcolor{revise_text}{The comparison results on two real-world noisy datasets, CIFAR-10N and CIFAR-100N, are shown in Table~\ref{tab: comparison_real_noise}.}
Except for GLS and Ours$^*$, other methods use the same setting of hyper-parameters for all noise rates on each dataset.
It reveals that our ILDE outperforms the best competitor by 2.77\%/1.31\% on CIFAR-100N/CIFAR-10N-Worst (p-value=6.52e-5/9.68e-3$<$0.05). 
\textcolor{revise_text}{Moreover, we also extended our comparisons to include Clothing1M, a large-scale real-world noisy dataset. The performance score for TNLPAD was extracted directly from its original publication~\cite{wang2024tackling}, and we have ensured that our ILDE method and other competing methods were evaluated under identical conditions as reported in~\cite{wang2024tackling} to maintain comparability. 
The results, as presented in Table~\ref{tab: comparison_Clothing1M}, demonstrate that our ILDE method outperforms TNLPAD, the best-performing competitor, by a clear margin (72.45\% vs. 71.51\%).}
The results suggest that the proposed ILDE can help the trained classifier combat real-world label noise, which is instance-dependent and more challenging than synthetic label noise.

\begin{table}[t]
\setlength\tabcolsep{8.0pt}
\begin{center}
\caption{Test accuracy (\%, mean$\pm$standard deviation) of ILDE and its variants on CIFAR-10 with asymmetric noise (noise rate=0.5). The best result is highlighted with \textbf{bold}.}
\label{tab: ablation_LDR}
\begin{tabular}{ccc|c}
\hline \hline
$\mathcal{L}_{NL}$ & $\mathcal{L}_{LD}$ & $\mathcal{L}_M$ & Accuracy \\ \hline
$\checkmark$     &                  &                   & 46.81 $\pm$ 0.57   \\
$\checkmark$     & $\checkmark$     &                   & 70.19 $\pm$ 0.56 \\
$\checkmark$     & $\checkmark$     & $\checkmark$      & \textbf{80.74 $\pm$ 0.51} \\ \hline \hline
\end{tabular}
\end{center}
\end{table}

\begin{table}[t]
\begin{center}
\caption{Test accuracy (\%, mean$\pm$standard deviation) of the proposed ILDE with complete $L_{LD}$ and its variants on CIFAR-10 with asymmetric noise (noise rate=0.5). The best result is highlighted with \textbf{bold}.}
\label{tab: analysis_LD}
\begin{tabular}{cc|c}
\hline \hline
Temporal Ensembling & Sharpening    & Accuracy \\ \hline
                    &               & 75.85 $\pm$ 0.60 \\
$\checkmark$        &               & 79.71 $\pm$ 0.30 \\
$\checkmark$        & $\checkmark$  & \textbf{80.74 $\pm$ 0.51} \\ \hline \hline
\end{tabular}
\end{center}
\end{table}

\subsection{Ablation Studies}
We conducted ablation studies on CIFAR-10 with asymmetric noise (noise rate is 0.5) to investigate the effectiveness of $\mathcal{L}_{LD}$ and $\mathcal{L}_M$ in ILDE.
We validated ILDE and its two variants and displayed the results in Table~\ref{tab: ablation_LDR}, where
`$\mathcal{L}_{NL}$' means only using the cross-entropy loss between the predictions and noisy labels, 
`$\mathcal{L}_{NL}+\mathcal{L}_{LD}$' means using ILDE without diagonally-dominant constraint $\mathcal{L}_M$, and
`$\mathcal{L}_{NL}+\mathcal{L}_{LD}+\mathcal{L}_M$ represents the ILDE.
It shows that the performance of ILDE drops seriously when $\mathcal{L}_M$ or both $\mathcal{L}_{LD}$ and $L_M$ are removed.

Moreover, there are two important operations in $\mathcal{L}_{LD}$ of ILDE, $i.e.$, the estimated instance-dependent LD \textit{sharpening} (Eq.~\ref{eq: tau}) and \textit{temporal ensembling} (Eq.~\ref{eq: beta}).
We conducted experiments on CIFAR-10 to validate their effectiveness.
Table~\ref{tab: analysis_LD} gives the performance of the ILDE with complete $\mathcal{L}_{LD}$ and its two variants.
It reveals that the performance of ILDE is degraded more or less when \textit{sharpening} or \textit{temporal ensembling} is removed.

\begin{figure*}[t]
\centering
\includegraphics[width=2.0\columnwidth]{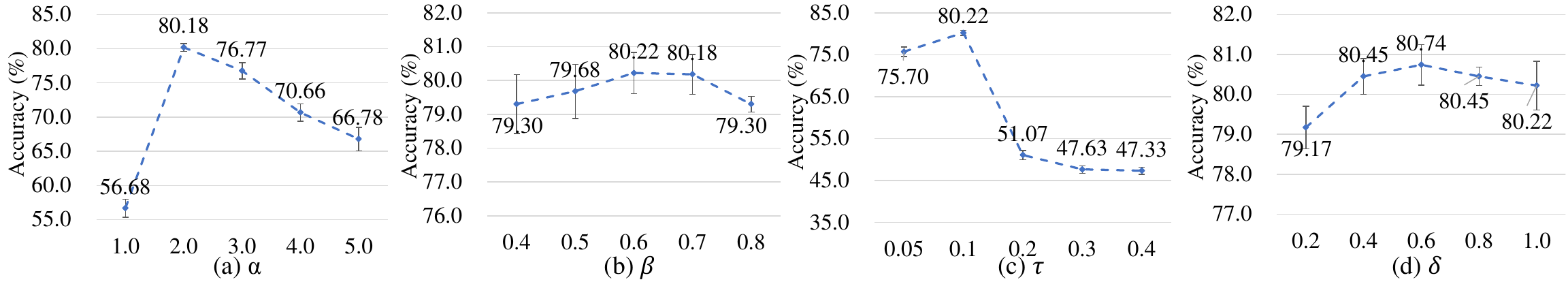}
\caption{
Test Accuracy (\%) of ILDE versus the value of (a) $\alpha$, (b) $\beta$, (c) $\tau$, and (d) $\delta$ on CIFAR-10 (asymmetric noise rate is 0.5).
The mean and standard deviation of the five runs' performances are shown here.
}
\label{fig: hyper_params}
\end{figure*}

\section{Discussions}

\subsection{Hyper-parameter Settings}

In the proposed ILDE, there are four hyper-parameters that should be adjusted, $i.e.$, the weighting factors $\alpha$ and $\delta$ in the total loss function (Eq.~\ref{eq: alpha_delta}), the weighting factor $\beta$ in temporal ensembling (Eq.~\ref{eq: beta}), and the sharpen factor $\tau$ in Eq.~\ref{eq: tau}.
We chose the CIFAR-10 dataset with an asymmetric noise rate of 0.5 for a case study. For the sake of computational complexity, we initialized these hyper-parameters as $\alpha=2.0, \beta=0.7, \tau=0.1, \delta=1.0$ and adjusted them one by one while fixing the other three.
The plot of model performance versus the value of each hyper-parameter was depicted in Fig.~\ref{fig: hyper_params}.
It shows that our ILDE method achieves the best performance in this case study when setting these hyper-parameters as $\alpha=2.0, \beta=0.6, \tau=0.1, \delta=0.6$.

\begin{table}[t]
\begin{center}
\setlength\tabcolsep{15.0pt}
\caption{Test accuracy (\%, mean$\pm$standard deviation) of ILDE trained on CIFAR-10 (asymmetric noise rate is 0.5) with different batch sizes. The best result is highlighted with \textbf{bold}.}
\label{tab: batch_size}
\begin{tabular}{c|c}
\hline \hline
Batch Size & Test Accuracy         \\ \hline
128        & 80.74 $\pm $ 0.51          \\
64         & 82.74 $\pm $ 0.55          \\
32         & \textbf{84.43 $\pm $ 0.43} \\
16         & 83.33 $\pm $ 0.65          \\
8          & 78.83 $\pm $ 0.69          \\ \hline \hline
\end{tabular}
\end{center}
\end{table}

\subsection{Analysis of Batch Size}

The noise transition matrix estimation is essential to the estimation of instance-dependent LD.
In our ILDE method, the noise transition matrix is estimated using Eq.~\ref{eq: noise transition matrix}, in which the batch size plays a pivotal role in the estimation.
We chose CIFAR-10 with an asymmetric noise rate of 0.5 as a case study to explore how the batch size affects the performance of our ILDE.
In this experiment, the hyper-parameters are set as $\alpha=2.0, \beta=0.6, \tau=0.1, \delta=0.6$.
Table~\ref{tab: batch_size} gives the test accuracy obtained by our ILDE with different settings of the batch size.
It shows that the accuracy of ILDE improves from 80.74\% to 84.43\% when the batch size drops from 128 to 32. It can be attributed to the fact that, when the batch size gets smaller, the estimated noise transition matrix becomes more suitable for each sample in the mini-batch.
However, when the batch size continues to drop from 32 to 8, the performance of ILDE deteriorates. 
The reason is that a batch of samples that is too small is insufficient to accurately characterize the inter-class correlation.

\begin{figure}[t]
\centering
\includegraphics[width=1.0\columnwidth]{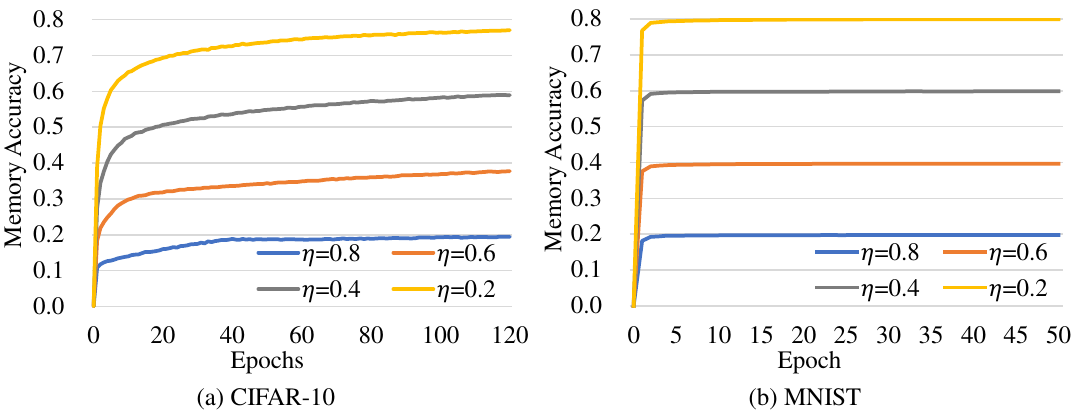}
\caption{
Memory accuracy of classifiers trained on MNIST and CIFAR-10 datasets with symmetric noise ($noise\ rate \in [0.2, 0.4, 0.6, 0.8]$).
Memory accuracy is calculated based on the predictions and noisy labels of training data. 
}
\label{fig: memory_acc}
\end{figure}

\begin{figure*}[t]
\centering
\includegraphics[width=2.0\columnwidth]{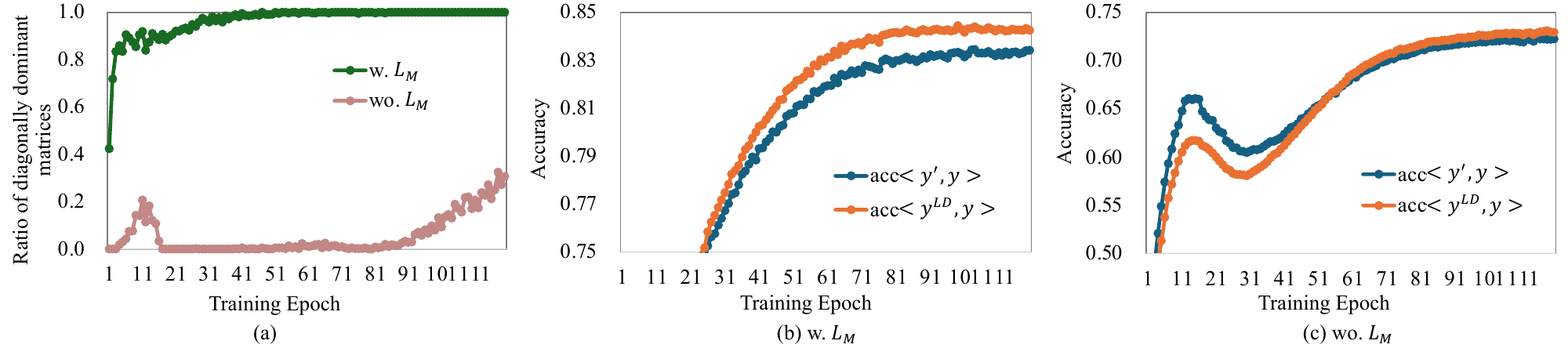}
\caption{
\textcolor{revise_text}{Influence of diagonally dominant regularization $L_M$ on the accuracy of corrected labels when validating our ILDE on CIFAR10 with asymmetric noise and noise rate equals 0.5. 
Three sub-figures demonstrate (a) the impact of $L_M$ on the ratio of estimated noise transition matrices that are diagonally dominant over epochs, (b, c) the accuracy of model predictions $\textbf{\textit{y}}^{'}$ and corrected labels $\textbf{\textit{y}}^{LD}$ .}
}
\label{fig: L_M}
\end{figure*}

\subsection{Memorizing Noisy Training Data}
We visualized the memory accuracy curves of the proposed ILDE in Fig.~\ref{fig: memory_acc}. The model was trained on the synthetic noisy datasets MNIST and CIFAR-10, whose symmetric noise rates are set to $[0.2, 0.4, 0.6, 0.8]$, respectively.
The memory accuracy is calculated based on the predictions and noisy labels of training data.
It is expected that the classifier memorizes all training samples with clean labels, and does not memorize those with noisy labels.
It shows that the memory accuracy curves on both datasets converge to 0.2 in the hardest case, where the noise rate is 0.8, indicating that the proposed ILDE can effectively prevent the classifier from memorizing noisy labels.

\subsection{Analysis of $L_M$}
\textcolor{revise_text}{
Our approach is predicated on the clean-labels-domination assumption~\cite{wei2021learning}, which posits that a training sample is more likely to be correctly labeled than mislabeled. 
This assumption is crucial for the noise transition matrix to maintain the integrity of class label semantics by being diagonally dominant. 
However, the noise transition matrix estimated from batch-level predictions via the Gram matrix may not inherently satisfy this condition. 
To ensure that the matrix is diagonally dominant without being identity, we introduce the regularization $L_M$ and meticulously tune its weight factor $\delta$.}

\textcolor{revise_text}{
Fig.~\ref{fig: L_M}(a) illustrates that, with $L_M$, the estimated noise transition matrices gradually achieve diagonal dominance, which is crucial for correcting a portion of samples mislabeled by the network via $\textbf{\textit{y}}^{LD}=\textbf{\textit{y}}^{'} \times M^{LD^T}$. 
As depicted in Fig.~\ref{fig: L_M}(b), the accuracy of the corrected labels $\textbf{\textit{y}}^{LD}$ is consistently higher than that of the network prediction $\textbf{\textit{y}}^{'}$, and the comparison with “training without $L_M$”, as shown in Fig.~\ref{fig: L_M}(c), further confirm that diagonally dominant noise transition matrices effectively transform noisy posterior probabilities into clean ones, thus alleviating the impact of label noise over the training process.}

\begin{table}[t]
\begin{center}
\setlength\tabcolsep{15.0pt}
\caption{Training time cost of our ILDE and all competing methods on CIFAR-10N.}
\label{tab: training_time}
\begin{tabular}{l|c}
\hline \hline
Method   & Training Time Cost \\ \hline
Baseline & 18min17s      \\ 
SCE      & 18min36s      \\ 
ELR      & 19min19s      \\ 
NCE+AUL  & 19min33s      \\ 
NCE+AGCE & 19min32s      \\ 
SR       & 20min07s      \\ 
GLS      & 18min48s      \\
VolMinNet & 18min45s     \\
MEIDTM   & 18min09s      \\
Our ILDE  & 18min39s      \\ \hline \hline
\end{tabular}
\end{center}
\end{table}

\subsection{Training Time Cost}
Table~\ref{tab: training_time} shows the training time cost of our ILDE and all competing methods on the CIFAR-10N dataset.
It shows that the training time of our ILDE is comparable with that of Baseline, SCE, VolMinNet, MEIDTM, and GLS.

\section{Conclusion}
\textcolor{revise_text}{
In this study, we introduce the ILDE method for training DCNNs on datasets with noisy labels for image classification. 
ILDE estimates the instance-dependent LD labels and uses them as supervision to avert the DCNN's memorization of noisy labels. 
Unlike traditional noise transition matrix approaches, ILDE approximates the instance-dependent noise transition matrix by assessing inter-class correlations in mini-batch samples, eliminating the need for (pseudo) anchor points and simplifying the process to a single training phase. 
Our evaluation of ILDE against eight other methods on synthetic and real-world noisy datasets demonstrates superior performance, with ablation studies confirming the effectiveness of ILDE's components.}

\noindent \textcolor{revise_text}{\textbf{Limitation} While the ILDE method is effective, it's sensitive to batch size and the diversity of features within each mini-batch. Ensuring a mini-batch that is representative of all classes is essential for the accurate estimation of the noise transition matrix. We are committed to addressing this limitation in our future work to improve ILDE's adaptability and robustness, making it a more versatile and reliable tool for tackling label noise in a wide array of vision tasks.
}

\backmatter
\bmhead{Acknowledgments}
This work was supported in part by the National Natural Science Foundation of China under Grant 62171377, in part by the Natural Science Foundation of Ningbo City, China, under Grant 2021J052, and in part by the Innovation Foundation for Doctor Dissertation of Northwestern Polytechnical University under Grant CX2022056.

\bmhead{Data availability statements}
The findings of this study are supported by the MNIST dataset, CIFAR-10 dataset, and CIFAR-N datasets. 
The MNIST dataset is openly available at \url{http://yann.lecun.com/exdb/mnist/}. 
The CIFAR-10 dataset is available at \url{https://www.cs.toronto.edu/~kriz/cifar.html}. 
The CIFAR-N datasets are available at \url{http://noisylabels.com/}.
\textcolor{revise_text}{The Clothing1M dataset is available at \url{https://www.v7labs.com/open-datasets/clothing1m}.}

\bibliography{sn-bibliography}


\begin{thebibliography}{55}
\ifx \bisbn   \undefined \def \bisbn  #1{ISBN #1}\fi
\ifx \binits  \undefined \def \binits#1{#1}\fi
\ifx \bauthor  \undefined \def \bauthor#1{#1}\fi
\ifx \batitle  \undefined \def \batitle#1{#1}\fi
\ifx \bjtitle  \undefined \def \bjtitle#1{#1}\fi
\ifx \bvolume  \undefined \def \bvolume#1{\textbf{#1}}\fi
\ifx \byear  \undefined \def \byear#1{#1}\fi
\ifx \bissue  \undefined \def \bissue#1{#1}\fi
\ifx \bfpage  \undefined \def \bfpage#1{#1}\fi
\ifx \blpage  \undefined \def \blpage #1{#1}\fi
\ifx \burl  \undefined \def \burl#1{\textsf{#1}}\fi
\ifx \doiurl  \undefined \def \doiurl#1{\url{https://doi.org/#1}}\fi
\ifx \betal  \undefined \def \betal{\textit{et al.}}\fi
\ifx \binstitute  \undefined \def \binstitute#1{#1}\fi
\ifx \binstitutionaled  \undefined \def \binstitutionaled#1{#1}\fi
\ifx \bctitle  \undefined \def \bctitle#1{#1}\fi
\ifx \beditor  \undefined \def \beditor#1{#1}\fi
\ifx \bpublisher  \undefined \def \bpublisher#1{#1}\fi
\ifx \bbtitle  \undefined \def \bbtitle#1{#1}\fi
\ifx \bedition  \undefined \def \bedition#1{#1}\fi
\ifx \bseriesno  \undefined \def \bseriesno#1{#1}\fi
\ifx \blocation  \undefined \def \blocation#1{#1}\fi
\ifx \bsertitle  \undefined \def \bsertitle#1{#1}\fi
\ifx \bsnm \undefined \def \bsnm#1{#1}\fi
\ifx \bsuffix \undefined \def \bsuffix#1{#1}\fi
\ifx \bparticle \undefined \def \bparticle#1{#1}\fi
\ifx \barticle \undefined \def \barticle#1{#1}\fi
\bibcommenthead
\ifx \bconfdate \undefined \def \bconfdate #1{#1}\fi
\ifx \botherref \undefined \def \botherref #1{#1}\fi
\ifx \url \undefined \def \url#1{\textsf{#1}}\fi
\ifx \bchapter \undefined \def \bchapter#1{#1}\fi
\ifx \bbook \undefined \def \bbook#1{#1}\fi
\ifx \bcomment \undefined \def \bcomment#1{#1}\fi
\ifx \oauthor \undefined \def \oauthor#1{#1}\fi
\ifx \citeauthoryear \undefined \def \citeauthoryear#1{#1}\fi
\ifx \endbibitem  \undefined \def \endbibitem {}\fi
\ifx \bconflocation  \undefined \def \bconflocation#1{#1}\fi
\ifx \arxivurl  \undefined \def \arxivurl#1{\textsf{#1}}\fi
\csname PreBibitemsHook\endcsname

%%% 1
\bibitem{yu2018learning}
\begin{bchapter}
\bauthor{\bsnm{Yu}, \binits{X.}},
\bauthor{\bsnm{Liu}, \binits{T.}},
\bauthor{\bsnm{Gong}, \binits{M.}},
\bauthor{\bsnm{Tao}, \binits{D.}}:
\bctitle{Learning with biased complementary labels}.
In: \bbtitle{Proceedings of the European Conference on Computer Vision},
pp. \bfpage{68}--\blpage{83}
(\byear{2018})
\end{bchapter}
\endbibitem

%%% 2
\bibitem{karimi2020deep}
\begin{barticle}
\bauthor{\bsnm{Karimi}, \binits{D.}},
\bauthor{\bsnm{Dou}, \binits{H.}},
\bauthor{\bsnm{Warfield}, \binits{S.K.}},
\bauthor{\bsnm{Gholipour}, \binits{A.}}:
\batitle{Deep learning with noisy labels: Exploring techniques and remedies in
  medical image analysis}.
\bjtitle{Medical image analysis}
\bvolume{65},
\bfpage{101759}
(\byear{2020})
\end{barticle}
\endbibitem

%%% 3
\bibitem{algan2021image}
\begin{barticle}
\bauthor{\bsnm{Algan}, \binits{G.}},
\bauthor{\bsnm{Ulusoy}, \binits{I.}}:
\batitle{Image classification with deep learning in the presence of noisy
  labels: A survey}.
\bjtitle{Knowledge-Based Systems}
\bvolume{215},
\bfpage{106771}
(\byear{2021})
\end{barticle}
\endbibitem

%%% 4
\bibitem{song2022learning}
\begin{botherref}
\oauthor{\bsnm{Song}, \binits{H.}},
\oauthor{\bsnm{Kim}, \binits{M.}},
\oauthor{\bsnm{Park}, \binits{D.}},
\oauthor{\bsnm{Shin}, \binits{Y.}},
\oauthor{\bsnm{Lee}, \binits{J.-G.}}:
Learning from noisy labels with deep neural networks: A survey.
IEEE Transactions on Neural Networks and Learning Systems
(2022)
\end{botherref}
\endbibitem

%%% 5
\bibitem{Zhang2017UnderstandingDL}
\begin{botherref}
\oauthor{\bsnm{Zhang}, \binits{C.}},
\oauthor{\bsnm{Bengio}, \binits{S.}},
\oauthor{\bsnm{Hardt}, \binits{M.}},
\oauthor{\bsnm{Recht}, \binits{B.}},
\oauthor{\bsnm{Vinyals}, \binits{O.}}:
Understanding deep learning requires rethinking generalization.
International Conference on Learning Representations
(2017)
\end{botherref}
\endbibitem

%%% 6
\bibitem{han2018co}
\begin{botherref}
\oauthor{\bsnm{Han}, \binits{B.}},
\oauthor{\bsnm{Yao}, \binits{Q.}},
\oauthor{\bsnm{Yu}, \binits{X.}},
\oauthor{\bsnm{Niu}, \binits{G.}},
\oauthor{\bsnm{Xu}, \binits{M.}},
\oauthor{\bsnm{Hu}, \binits{W.}},
\oauthor{\bsnm{Tsang}, \binits{I.}},
\oauthor{\bsnm{Sugiyama}, \binits{M.}}:
Co-teaching: Robust training of deep neural networks with extremely noisy
  labels.
Advances in Neural Information Processing Systems
\textbf{31}
(2018)
\end{botherref}
\endbibitem

%%% 7
\bibitem{li2019dividemix}
\begin{bchapter}
\bauthor{\bsnm{Li}, \binits{J.}},
\bauthor{\bsnm{Socher}, \binits{R.}},
\bauthor{\bsnm{Hoi}, \binits{S.C.}}:
\bctitle{Dividemix: Learning with noisy labels as semi-supervised learning}.
In: \bbtitle{International Conference on Learning Representations}
(\byear{2019})
\end{bchapter}
\endbibitem

%%% 8
\bibitem{wang2022scalable}
\begin{bchapter}
\bauthor{\bsnm{Wang}, \binits{Y.}},
\bauthor{\bsnm{Sun}, \binits{X.}},
\bauthor{\bsnm{Fu}, \binits{Y.}}:
\bctitle{Scalable penalized regression for noise detection in learning with
  noisy labels}.
In: \bbtitle{Proceedings of the IEEE/CVF Conference on Computer Vision and
  Pattern Recognition},
pp. \bfpage{346}--\blpage{355}
(\byear{2022})
\end{bchapter}
\endbibitem

%%% 9
\bibitem{xia2020robust}
\begin{bchapter}
\bauthor{\bsnm{Xia}, \binits{X.}},
\bauthor{\bsnm{Liu}, \binits{T.}},
\bauthor{\bsnm{Han}, \binits{B.}},
\bauthor{\bsnm{Gong}, \binits{C.}},
\bauthor{\bsnm{Wang}, \binits{N.}},
\bauthor{\bsnm{Ge}, \binits{Z.}},
\bauthor{\bsnm{Chang}, \binits{Y.}}:
\bctitle{Robust early-learning: Hindering the memorization of noisy labels}.
In: \bbtitle{International Conference on Learning Representations}
(\byear{2020})
\end{bchapter}
\endbibitem

%%% 10
\bibitem{kim2019nlnl}
\begin{bchapter}
\bauthor{\bsnm{Kim}, \binits{Y.}},
\bauthor{\bsnm{Yim}, \binits{J.}},
\bauthor{\bsnm{Yun}, \binits{J.}},
\bauthor{\bsnm{Kim}, \binits{J.}}:
\bctitle{Nlnl: Negative learning for noisy labels}.
In: \bbtitle{Proceedings of the IEEE/CVF International Conference on Computer
  Vision},
pp. \bfpage{101}--\blpage{110}
(\byear{2019})
\end{bchapter}
\endbibitem

%%% 11
\bibitem{thulasidasan2019combating}
\begin{bchapter}
\bauthor{\bsnm{Thulasidasan}, \binits{S.}},
\bauthor{\bsnm{Bhattacharya}, \binits{T.}},
\bauthor{\bsnm{Bilmes}, \binits{J.A.}},
\bauthor{\bsnm{Chennupati}, \binits{G.}},
\bauthor{\bsnm{Mohd-Yusof}, \binits{J.}}:
\bctitle{Combating label noise in deep learning using abstention}.
In: \bbtitle{International Conference on Machine Learning}
(\byear{2019})
\end{bchapter}
\endbibitem

%%% 12
\bibitem{patrini2017making}
\begin{bchapter}
\bauthor{\bsnm{Patrini}, \binits{G.}},
\bauthor{\bsnm{Rozza}, \binits{A.}},
\bauthor{\bsnm{Krishna~Menon}, \binits{A.}},
\bauthor{\bsnm{Nock}, \binits{R.}},
\bauthor{\bsnm{Qu}, \binits{L.}}:
\bctitle{Making deep neural networks robust to label noise: A loss correction
  approach}.
In: \bbtitle{Proceedings of the IEEE/CVF Conference on Computer Vision and
  Pattern Recognition},
pp. \bfpage{1944}--\blpage{1952}
(\byear{2017})
\end{bchapter}
\endbibitem

%%% 13
\bibitem{xia2022sample}
\begin{botherref}
\oauthor{\bsnm{Xia}, \binits{X.}},
\oauthor{\bsnm{Liu}, \binits{T.}},
\oauthor{\bsnm{Han}, \binits{B.}},
\oauthor{\bsnm{Gong}, \binits{M.}},
\oauthor{\bsnm{Yu}, \binits{J.}},
\oauthor{\bsnm{Niu}, \binits{G.}},
\oauthor{\bsnm{Sugiyama}, \binits{M.}}:
Sample selection with uncertainty of losses for learning with noisy labels.
International Conference on Learning Representations
(2022)
\end{botherref}
\endbibitem

%%% 14
\bibitem{Huang2022UncertaintyAwareLA}
\begin{botherref}
\oauthor{\bsnm{Huang}, \binits{Y.}},
\oauthor{\bsnm{Bai}, \binits{B.}},
\oauthor{\bsnm{Zhao}, \binits{S.}},
\oauthor{\bsnm{Bai}, \binits{K.}},
\oauthor{\bsnm{Wang}, \binits{F.}}:
Uncertainty-aware learning against label noise on imbalanced datasets.
Proceedings of the AAAI Conference on Artificial Intelligence
(2022)
\end{botherref}
\endbibitem

%%% 15
\bibitem{li2021provably}
\begin{bchapter}
\bauthor{\bsnm{Li}, \binits{X.}},
\bauthor{\bsnm{Liu}, \binits{T.}},
\bauthor{\bsnm{Han}, \binits{B.}},
\bauthor{\bsnm{Niu}, \binits{G.}},
\bauthor{\bsnm{Sugiyama}, \binits{M.}}:
\bctitle{Provably end-to-end label-noise learning without anchor points}.
In: \bbtitle{International Conference on Machine Learning},
pp. \bfpage{6403}--\blpage{6413}
(\byear{2021}).
\bcomment{PMLR}
\end{bchapter}
\endbibitem

%%% 16
\bibitem{cheng2022instance}
\begin{bchapter}
\bauthor{\bsnm{Cheng}, \binits{D.}},
\bauthor{\bsnm{Liu}, \binits{T.}},
\bauthor{\bsnm{Ning}, \binits{Y.}},
\bauthor{\bsnm{Wang}, \binits{N.}},
\bauthor{\bsnm{Han}, \binits{B.}},
\bauthor{\bsnm{Niu}, \binits{G.}},
\bauthor{\bsnm{Gao}, \binits{X.}},
\bauthor{\bsnm{Sugiyama}, \binits{M.}}:
\bctitle{Instance-dependent label-noise learning with manifold-regularized
  transition matrix estimation}.
In: \bbtitle{Proceedings of the IEEE/CVF Conference on Computer Vision and
  Pattern Recognition},
pp. \bfpage{16630}--\blpage{16639}
(\byear{2022})
\end{bchapter}
\endbibitem

%%% 17
\bibitem{hendrycks2018using}
\begin{botherref}
\oauthor{\bsnm{Hendrycks}, \binits{D.}},
\oauthor{\bsnm{Mazeika}, \binits{M.}},
\oauthor{\bsnm{Wilson}, \binits{D.}},
\oauthor{\bsnm{Gimpel}, \binits{K.}}:
Using trusted data to train deep networks on labels corrupted by severe noise.
Advances in neural information processing systems
\textbf{31}
(2018)
\end{botherref}
\endbibitem

%%% 18
\bibitem{xia2019anchor}
\begin{botherref}
\oauthor{\bsnm{Xia}, \binits{X.}},
\oauthor{\bsnm{Liu}, \binits{T.}},
\oauthor{\bsnm{Wang}, \binits{N.}},
\oauthor{\bsnm{Han}, \binits{B.}},
\oauthor{\bsnm{Gong}, \binits{C.}},
\oauthor{\bsnm{Niu}, \binits{G.}},
\oauthor{\bsnm{Sugiyama}, \binits{M.}}:
Are anchor points really indispensable in label-noise learning?
Advances in Neural Information Processing Systems
\textbf{32}
(2019)
\end{botherref}
\endbibitem

%%% 19
\bibitem{shu2020meta}
\begin{botherref}
\oauthor{\bsnm{Shu}, \binits{J.}},
\oauthor{\bsnm{Zhao}, \binits{Q.}},
\oauthor{\bsnm{Xu}, \binits{Z.}},
\oauthor{\bsnm{Meng}, \binits{D.}}:
Meta transition adaptation for robust deep learning with noisy labels.
arXiv preprint arXiv:2006.05697
(2020)
\end{botherref}
\endbibitem

%%% 20
\bibitem{yao2020dual}
\begin{barticle}
\bauthor{\bsnm{Yao}, \binits{Y.}},
\bauthor{\bsnm{Liu}, \binits{T.}},
\bauthor{\bsnm{Han}, \binits{B.}},
\bauthor{\bsnm{Gong}, \binits{M.}},
\bauthor{\bsnm{Deng}, \binits{J.}},
\bauthor{\bsnm{Niu}, \binits{G.}},
\bauthor{\bsnm{Sugiyama}, \binits{M.}}:
\batitle{Dual t: Reducing estimation error for transition matrix in label-noise
  learning}.
\bjtitle{Advances in neural information processing systems}
\bvolume{33},
\bfpage{7260}--\blpage{7271}
(\byear{2020})
\end{barticle}
\endbibitem

%%% 21
\bibitem{cheng2020learning}
\begin{bchapter}
\bauthor{\bsnm{Cheng}, \binits{H.}},
\bauthor{\bsnm{Zhu}, \binits{Z.}},
\bauthor{\bsnm{Li}, \binits{X.}},
\bauthor{\bsnm{Gong}, \binits{Y.}},
\bauthor{\bsnm{Sun}, \binits{X.}},
\bauthor{\bsnm{Liu}, \binits{Y.}}:
\bctitle{Learning with instance-dependent label noise: A sample sieve
  approach}.
In: \bbtitle{International Conference on Learning Representations}
(\byear{2020})
\end{bchapter}
\endbibitem

%%% 22
\bibitem{berthon2021confidence}
\begin{bchapter}
\bauthor{\bsnm{Berthon}, \binits{A.}},
\bauthor{\bsnm{Han}, \binits{B.}},
\bauthor{\bsnm{Niu}, \binits{G.}},
\bauthor{\bsnm{Liu}, \binits{T.}},
\bauthor{\bsnm{Sugiyama}, \binits{M.}}:
\bctitle{Confidence scores make instance-dependent label-noise learning
  possible}.
In: \bbtitle{International Conference on Machine Learning},
pp. \bfpage{825}--\blpage{836}
(\byear{2021}).
\bcomment{PMLR}
\end{bchapter}
\endbibitem

%%% 23
\bibitem{xia2020part}
\begin{barticle}
\bauthor{\bsnm{Xia}, \binits{X.}},
\bauthor{\bsnm{Liu}, \binits{T.}},
\bauthor{\bsnm{Han}, \binits{B.}},
\bauthor{\bsnm{Wang}, \binits{N.}},
\bauthor{\bsnm{Gong}, \binits{M.}},
\bauthor{\bsnm{Liu}, \binits{H.}},
\bauthor{\bsnm{Niu}, \binits{G.}},
\bauthor{\bsnm{Tao}, \binits{D.}},
\bauthor{\bsnm{Sugiyama}, \binits{M.}}:
\batitle{Part-dependent label noise: Towards instance-dependent label noise}.
\bjtitle{Advances in Neural Information Processing Systems}
\bvolume{33},
\bfpage{7597}--\blpage{7610}
(\byear{2020})
\end{barticle}
\endbibitem

%%% 24
\bibitem{yang2021estimating}
\begin{botherref}
\oauthor{\bsnm{Yang}, \binits{S.}},
\oauthor{\bsnm{Yang}, \binits{E.}},
\oauthor{\bsnm{Han}, \binits{B.}},
\oauthor{\bsnm{Liu}, \binits{Y.}},
\oauthor{\bsnm{Xu}, \binits{M.}},
\oauthor{\bsnm{Niu}, \binits{G.}},
\oauthor{\bsnm{Liu}, \binits{T.}}:
Estimating instance-dependent label-noise transition matrix using dnns.
arXiv preprint arXiv:2105.13001
(2021)
\end{botherref}
\endbibitem

%%% 25
\bibitem{kye2022learning}
\begin{bchapter}
\bauthor{\bsnm{Kye}, \binits{S.M.}},
\bauthor{\bsnm{Choi}, \binits{K.}},
\bauthor{\bsnm{Yi}, \binits{J.}},
\bauthor{\bsnm{Chang}, \binits{B.}}:
\bctitle{Learning with noisy labels by efficient transition matrix estimation
  to combat label miscorrection}.
In: \bbtitle{Proceedings of the European Conference on Computer Vision},
pp. \bfpage{717}--\blpage{738}
(\byear{2022}).
\bcomment{Springer}
\end{bchapter}
\endbibitem

%%% 26
\bibitem{wei2021learning}
\begin{bchapter}
\bauthor{\bsnm{Wei}, \binits{J.}},
\bauthor{\bsnm{Zhu}, \binits{Z.}},
\bauthor{\bsnm{Cheng}, \binits{H.}},
\bauthor{\bsnm{Liu}, \binits{T.}},
\bauthor{\bsnm{Niu}, \binits{G.}},
\bauthor{\bsnm{Liu}, \binits{Y.}}:
\bctitle{Learning with noisy labels revisited: A study using real-world human
  annotations}.
In: \bbtitle{International Conference on Learning Representations}
(\byear{2021})
\end{bchapter}
\endbibitem

%%% 27
\bibitem{cheng2022class}
\begin{barticle}
\bauthor{\bsnm{Cheng}, \binits{D.}},
\bauthor{\bsnm{Ning}, \binits{Y.}},
\bauthor{\bsnm{Wang}, \binits{N.}},
\bauthor{\bsnm{Gao}, \binits{X.}},
\bauthor{\bsnm{Yang}, \binits{H.}},
\bauthor{\bsnm{Du}, \binits{Y.}},
\bauthor{\bsnm{Han}, \binits{B.}},
\bauthor{\bsnm{Liu}, \binits{T.}}:
\batitle{Class-dependent label-noise learning with cycle-consistency
  regularization}.
\bjtitle{Advances in Neural Information Processing Systems}
\bvolume{35},
\bfpage{11104}--\blpage{11116}
(\byear{2022})
\end{barticle}
\endbibitem

%%% 28
\bibitem{liu2020early}
\begin{barticle}
\bauthor{\bsnm{Liu}, \binits{S.}},
\bauthor{\bsnm{Niles-Weed}, \binits{J.}},
\bauthor{\bsnm{Razavian}, \binits{N.}},
\bauthor{\bsnm{Fernandez-Granda}, \binits{C.}}:
\batitle{Early-learning regularization prevents memorization of noisy labels}.
\bjtitle{Advances in Neural Information Processing Systems}
\bvolume{33},
\bfpage{20331}--\blpage{20342}
(\byear{2020})
\end{barticle}
\endbibitem

%%% 29
\bibitem{Laine2017TemporalEnsembling}
\begin{bchapter}
\bauthor{\bsnm{Laine}, \binits{S.}},
\bauthor{\bsnm{Aila}, \binits{T.}}:
\bctitle{Temporal ensembling for semi-supervised learning}.
In: \bbtitle{International Conference on Learning Representations}
(\byear{2017})
\end{bchapter}
\endbibitem

%%% 30
\bibitem{shen2019learning}
\begin{bchapter}
\bauthor{\bsnm{Shen}, \binits{Y.}},
\bauthor{\bsnm{Sanghavi}, \binits{S.}}:
\bctitle{Learning with bad training data via iterative trimmed loss
  minimization}.
In: \bbtitle{International Conference on Machine Learning},
pp. \bfpage{5739}--\blpage{5748}
(\byear{2019})
\end{bchapter}
\endbibitem

%%% 31
\bibitem{nguyen2019self}
\begin{bchapter}
\bauthor{\bsnm{Nguyen}, \binits{D.T.}},
\bauthor{\bsnm{Mummadi}, \binits{C.K.}},
\bauthor{\bsnm{Ngo}, \binits{T.P.N.}},
\bauthor{\bsnm{Nguyen}, \binits{T.H.P.}},
\bauthor{\bsnm{Beggel}, \binits{L.}},
\bauthor{\bsnm{Brox}, \binits{T.}}:
\bctitle{Self: Learning to filter noisy labels with self-ensembling}.
In: \bbtitle{International Conference on Learning Representations}
(\byear{2019})
\end{bchapter}
\endbibitem

%%% 32
\bibitem{jiang2018mentornet}
\begin{bchapter}
\bauthor{\bsnm{Jiang}, \binits{L.}},
\bauthor{\bsnm{Zhou}, \binits{Z.}},
\bauthor{\bsnm{Leung}, \binits{T.}},
\bauthor{\bsnm{Li}, \binits{L.-J.}},
\bauthor{\bsnm{Fei-Fei}, \binits{L.}}:
\bctitle{Mentornet: Learning data-driven curriculum for very deep neural
  networks on corrupted labels}.
In: \bbtitle{International Conference on Machine Learning},
pp. \bfpage{2304}--\blpage{2313}
(\byear{2018})
\end{bchapter}
\endbibitem

%%% 33
\bibitem{Wei2022ToSO}
\begin{bchapter}
\bauthor{\bsnm{Wei}, \binits{J.}},
\bauthor{\bsnm{Liu}, \binits{H.}},
\bauthor{\bsnm{Liu}, \binits{T.}},
\bauthor{\bsnm{Niu}, \binits{G.}},
\bauthor{\bsnm{Liu}, \binits{Y.}}:
\bctitle{To smooth or not? when label smoothing meets noisy labels}.
In: \bbtitle{International Conference on Machine Learning}
(\byear{2022})
\end{bchapter}
\endbibitem

%%% 34
\bibitem{sukhbaatar2015training}
\begin{bchapter}
\bauthor{\bsnm{Sukhbaatar}, \binits{S.}},
\bauthor{\bsnm{Bruna}, \binits{J.}},
\bauthor{\bsnm{Paluri}, \binits{M.}},
\bauthor{\bsnm{Bourdev}, \binits{L.}},
\bauthor{\bsnm{Fergus}, \binits{R.}}:
\bctitle{Training convolutional networks with noisy labels}.
In: \bbtitle{International Conference on Learning Representations}
(\byear{2015})
\end{bchapter}
\endbibitem

%%% 35
\bibitem{ghosh2017robust}
\begin{bchapter}
\bauthor{\bsnm{Ghosh}, \binits{A.}},
\bauthor{\bsnm{Kumar}, \binits{H.}},
\bauthor{\bsnm{Sastry}, \binits{P.S.}}:
\bctitle{Robust loss functions under label noise for deep neural networks}.
In: \bbtitle{Proceedings of the AAAI Conference on Artificial Intelligence},
vol. \bseriesno{31}
(\byear{2017})
\end{bchapter}
\endbibitem

%%% 36
\bibitem{zhang2018generalized}
\begin{botherref}
\oauthor{\bsnm{Zhang}, \binits{Z.}},
\oauthor{\bsnm{Sabuncu}, \binits{M.}}:
Generalized cross entropy loss for training deep neural networks with noisy
  labels.
Advances in Neural Information Processing Systems
\textbf{31}
(2018)
\end{botherref}
\endbibitem

%%% 37
\bibitem{wang2019symmetric}
\begin{bchapter}
\bauthor{\bsnm{Wang}, \binits{Y.}},
\bauthor{\bsnm{Ma}, \binits{X.}},
\bauthor{\bsnm{Chen}, \binits{Z.}},
\bauthor{\bsnm{Luo}, \binits{Y.}},
\bauthor{\bsnm{Yi}, \binits{J.}},
\bauthor{\bsnm{Bailey}, \binits{J.}}:
\bctitle{Symmetric cross entropy for robust learning with noisy labels}.
In: \bbtitle{Proceedings of the IEEE/CVF International Conference on Computer
  Vision},
pp. \bfpage{322}--\blpage{330}
(\byear{2019})
\end{bchapter}
\endbibitem

%%% 38
\bibitem{lyu2019curriculum}
\begin{botherref}
\oauthor{\bsnm{Lyu}, \binits{Y.}},
\oauthor{\bsnm{Tsang}, \binits{I.W.}}:
Curriculum loss: Robust learning and generalization against label corruption.
International Conference on Learning Representations
(2019)
\end{botherref}
\endbibitem

%%% 39
\bibitem{ma2020normalized}
\begin{bchapter}
\bauthor{\bsnm{Ma}, \binits{X.}},
\bauthor{\bsnm{Huang}, \binits{H.}},
\bauthor{\bsnm{Wang}, \binits{Y.}},
\bauthor{\bsnm{Romano}, \binits{S.}},
\bauthor{\bsnm{Erfani}, \binits{S.}},
\bauthor{\bsnm{Bailey}, \binits{J.}}:
\bctitle{Normalized loss functions for deep learning with noisy labels}.
In: \bbtitle{International Conference on Machine Learning},
pp. \bfpage{6543}--\blpage{6553}
(\byear{2020})
\end{bchapter}
\endbibitem

%%% 40
\bibitem{yao2020deep}
\begin{bchapter}
\bauthor{\bsnm{Yao}, \binits{Y.}},
\bauthor{\bsnm{Deng}, \binits{J.}},
\bauthor{\bsnm{Chen}, \binits{X.}},
\bauthor{\bsnm{Gong}, \binits{C.}},
\bauthor{\bsnm{Wu}, \binits{J.}},
\bauthor{\bsnm{Yang}, \binits{J.}}:
\bctitle{Deep discriminative cnn with temporal ensembling for
  ambiguously-labeled image classification}.
In: \bbtitle{Proceedings of the AAAI Conference on Artificial Intelligence},
vol. \bseriesno{34},
pp. \bfpage{12669}--\blpage{12676}
(\byear{2020})
\end{bchapter}
\endbibitem

%%% 41
\bibitem{zhou2021asymmetric}
\begin{bchapter}
\bauthor{\bsnm{Zhou}, \binits{X.}},
\bauthor{\bsnm{Liu}, \binits{X.}},
\bauthor{\bsnm{Jiang}, \binits{J.}},
\bauthor{\bsnm{Gao}, \binits{X.}},
\bauthor{\bsnm{Ji}, \binits{X.}}:
\bctitle{Asymmetric loss functions for learning with noisy labels}.
In: \bbtitle{International Conference on Machine Learning},
pp. \bfpage{12846}--\blpage{12856}
(\byear{2021})
\end{bchapter}
\endbibitem

%%% 42
\bibitem{menon2019can}
\begin{bchapter}
\bauthor{\bsnm{Menon}, \binits{A.K.}},
\bauthor{\bsnm{Rawat}, \binits{A.S.}},
\bauthor{\bsnm{Reddi}, \binits{S.J.}},
\bauthor{\bsnm{Kumar}, \binits{S.}}:
\bctitle{Can gradient clipping mitigate label noise?}
In: \bbtitle{International Conference on Learning Representations}
(\byear{2019})
\end{bchapter}
\endbibitem

%%% 43
\bibitem{zhou2021learning}
\begin{bchapter}
\bauthor{\bsnm{Zhou}, \binits{X.}},
\bauthor{\bsnm{Liu}, \binits{X.}},
\bauthor{\bsnm{Wang}, \binits{C.}},
\bauthor{\bsnm{Zhai}, \binits{D.}},
\bauthor{\bsnm{Jiang}, \binits{J.}},
\bauthor{\bsnm{Ji}, \binits{X.}}:
\bctitle{Learning with noisy labels via sparse regularization}.
In: \bbtitle{Proceedings of the IEEE/CVF International Conference on Computer
  Vision},
pp. \bfpage{72}--\blpage{81}
(\byear{2021})
\end{bchapter}
\endbibitem

%%% 44
\bibitem{natarajan2013learning}
\begin{botherref}
\oauthor{\bsnm{Natarajan}, \binits{N.}},
\oauthor{\bsnm{Dhillon}, \binits{I.S.}},
\oauthor{\bsnm{Ravikumar}, \binits{P.K.}},
\oauthor{\bsnm{Tewari}, \binits{A.}}:
Learning with noisy labels.
Advances in neural information processing systems
\textbf{26}
(2013)
\end{botherref}
\endbibitem

%%% 45
\bibitem{yu2018efficient}
\begin{bchapter}
\bauthor{\bsnm{Yu}, \binits{X.}},
\bauthor{\bsnm{Liu}, \binits{T.}},
\bauthor{\bsnm{Gong}, \binits{M.}},
\bauthor{\bsnm{Batmanghelich}, \binits{K.}},
\bauthor{\bsnm{Tao}, \binits{D.}}:
\bctitle{An efficient and provable approach for mixture proportion estimation
  using linear independence assumption}.
In: \bbtitle{Proceedings of the IEEE Conference on Computer Vision and Pattern
  Recognition},
pp. \bfpage{4480}--\blpage{4489}
(\byear{2018})
\end{bchapter}
\endbibitem

%%% 46
\bibitem{yao2020towards}
\begin{botherref}
\oauthor{\bsnm{Yao}, \binits{Y.}},
\oauthor{\bsnm{Liu}, \binits{T.}},
\oauthor{\bsnm{Han}, \binits{B.}},
\oauthor{\bsnm{Gong}, \binits{M.}},
\oauthor{\bsnm{Niu}, \binits{G.}},
\oauthor{\bsnm{Sugiyama}, \binits{M.}},
\oauthor{\bsnm{Tao}, \binits{D.}}:
Towards mixture proportion estimation without irreducibility.
arXiv preprint arXiv:2002.03673
(2020)
\end{botherref}
\endbibitem

%%% 47
\bibitem{cohen2020separability}
\begin{barticle}
\bauthor{\bsnm{Cohen}, \binits{U.}},
\bauthor{\bsnm{Chung}, \binits{S.}},
\bauthor{\bsnm{Lee}, \binits{D.D.}},
\bauthor{\bsnm{Sompolinsky}, \binits{H.}}:
\batitle{Separability and geometry of object manifolds in deep neural
  networks}.
\bjtitle{Nature Communications}
\bvolume{11}(\bissue{1}),
\bfpage{1}--\blpage{13}
(\byear{2020})
\end{barticle}
\endbibitem

%%% 48
\bibitem{beyer2020we}
\begin{botherref}
\oauthor{\bsnm{Beyer}, \binits{L.}},
\oauthor{\bsnm{H{\'e}naff}, \binits{O.J.}},
\oauthor{\bsnm{Kolesnikov}, \binits{A.}},
\oauthor{\bsnm{Zhai}, \binits{X.}},
\oauthor{\bsnm{Oord}, \binits{A.v.d.}}:
Are we done with imagenet?
arXiv preprint arXiv:2006.07159
(2020)
\end{botherref}
\endbibitem

%%% 49
\bibitem{liu2020semi}
\begin{barticle}
\bauthor{\bsnm{Liu}, \binits{Q.}},
\bauthor{\bsnm{Yu}, \binits{L.}},
\bauthor{\bsnm{Luo}, \binits{L.}},
\bauthor{\bsnm{Dou}, \binits{Q.}},
\bauthor{\bsnm{Heng}, \binits{P.A.}}:
\batitle{Semi-supervised medical image classification with relation-driven
  self-ensembling model}.
\bjtitle{IEEE transactions on medical imaging}
\bvolume{39}(\bissue{11}),
\bfpage{3429}--\blpage{3440}
(\byear{2020})
\end{barticle}
\endbibitem

%%% 50
\bibitem{yang2022survey}
\begin{botherref}
\oauthor{\bsnm{Yang}, \binits{X.}},
\oauthor{\bsnm{Song}, \binits{Z.}},
\oauthor{\bsnm{King}, \binits{I.}},
\oauthor{\bsnm{Xu}, \binits{Z.}}:
A survey on deep semi-supervised learning.
IEEE Transactions on Knowledge and Data Engineering
(2022)
\end{botherref}
\endbibitem

%%% 51
\bibitem{lecun1998gradient}
\begin{barticle}
\bauthor{\bsnm{LeCun}, \binits{Y.}},
\bauthor{\bsnm{Bottou}, \binits{L.}},
\bauthor{\bsnm{Bengio}, \binits{Y.}},
\bauthor{\bsnm{Haffner}, \binits{P.}}:
\batitle{Gradient-based learning applied to document recognition}.
\bjtitle{Proceedings of the IEEE}
\bvolume{86}(\bissue{11}),
\bfpage{2278}--\blpage{2324}
(\byear{1998})
\end{barticle}
\endbibitem

%%% 52
\bibitem{krizhevsky2009learning}
\begin{botherref}
\oauthor{\bsnm{Krizhevsky}, \binits{A.}},
\oauthor{\bsnm{Hinton}, \binits{G.}}, et al.:
Learning multiple layers of features from tiny images
(2009)
\end{botherref}
\endbibitem

%%% 53
\bibitem{xiao2015learning}
\begin{bchapter}
\bauthor{\bsnm{Xiao}, \binits{T.}},
\bauthor{\bsnm{Xia}, \binits{T.}},
\bauthor{\bsnm{Yang}, \binits{Y.}},
\bauthor{\bsnm{Huang}, \binits{C.}},
\bauthor{\bsnm{Wang}, \binits{X.}}:
\bctitle{Learning from massive noisy labeled data for image classification}.
In: \bbtitle{Proceedings of the IEEE/CVF Conference on Computer Vision and
  Pattern Recognition},
pp. \bfpage{2691}--\blpage{2699}
(\byear{2015})
\end{bchapter}
\endbibitem

%%% 54
\bibitem{he2016deep}
\begin{bchapter}
\bauthor{\bsnm{He}, \binits{K.}},
\bauthor{\bsnm{Zhang}, \binits{X.}},
\bauthor{\bsnm{Ren}, \binits{S.}},
\bauthor{\bsnm{Sun}, \binits{J.}}:
\bctitle{Deep residual learning for image recognition}.
In: \bbtitle{Proceedings of the IEEE/CVF Conference on Computer Vision and
  Pattern Recognition},
pp. \bfpage{770}--\blpage{778}
(\byear{2016})
\end{bchapter}
\endbibitem

%%% 55
\bibitem{wang2024tackling}
\begin{botherref}
\oauthor{\bsnm{Wang}, \binits{J.}},
\oauthor{\bsnm{Xia}, \binits{X.}},
\oauthor{\bsnm{Lan}, \binits{L.}},
\oauthor{\bsnm{Wu}, \binits{X.}},
\oauthor{\bsnm{Yu}, \binits{J.}},
\oauthor{\bsnm{Yang}, \binits{W.}},
\oauthor{\bsnm{Han}, \binits{B.}},
\oauthor{\bsnm{Liu}, \binits{T.}}:
Tackling noisy labels with network parameter additive decomposition.
IEEE Transactions on Pattern Analysis and Machine Intelligence
(2024)
\end{botherref}
\endbibitem

\end{thebibliography}
\end{document}